\documentclass[sigconf]{acmart}

\copyrightyear{2026}
\acmYear{2026}
\setcopyright{cc}
\setcctype{by}
\acmConference[KDD 2026] {Proceedings of the 32nd ACM SIGKDD Conference on Knowledge Discovery and Data Mining V.1}{August 9--13, 2026}{Jeju Island, Republic of Korea.}
\acmBooktitle{Proceedings of the 32nd ACM SIGKDD Conference on Knowledge Discovery and Data Mining V.1 (KDD 2026), August 9--13, 2026, Jeju Island, Republic of Korea}
\acmISBN{979-8-4007-2258-5/2026/08}
\acmDOI{10.1145/3770854.3780318}

\AtBeginDocument{%
  \providecommand\BibTeX{{%
    \normalfont B\kern-0.5em{\scshape i\kern-0.25em b}\kern-0.8em\TeX}}}

\usepackage{algorithmic}
\usepackage{textcomp}
\usepackage{float}
\usepackage{diagbox}

\usepackage[utf8]{inputenc} 
\usepackage[T1]{fontenc}    
\usepackage{hyperref}       
\usepackage{url}            
\usepackage{booktabs}       
\usepackage{amsfonts}       
\usepackage{nicefrac}       
\usepackage{microtype}      
\usepackage{xcolor}         
\usepackage{graphicx}
\usepackage{subfigure}

\usepackage{amsmath}
 
\usepackage{amssymb}
\usepackage{mathtools}
\usepackage{amsthm}
\usepackage{multirow} 
\usepackage{makecell}
\usepackage{subcaption}
\usepackage{pifont}
\usepackage{enumitem}

\begin{document}


\title{From Tokenizer Bias to Backbone Capability: A Controlled Study of LLMs for Time Series Forecasting}

\author{Xinyu Zhang}
\affiliation{
  \institution{Harbin Institute of Technology}
  \city{Shenzhen}
  \country{China}
}
\affiliation{
  \institution{The Hong Kong Polytechnic University}
  \city{Hong Kong}
  \country{China}
}
\email{zhangxinyu45@stu.hit.edu.cn}

\author{Shanshan Feng}
\authornote{Corresponding author.}
\affiliation{%
  \institution{Wuhan University}
 \city{Wuhan}
  \country{China}}
\email{victor\_fengss@whu.edu.cn}

\author{Xutao Li}
\authornotemark[1]
\affiliation{%
 \institution{Harbin Institute of Technology}
 \city{Shenzhen}
 \country{China}}
  \email{lixutao@hit.edu.cn}

\author{Kenghong Lin}
\affiliation{%
 \institution{Harbin Institute of Technology}
 \city{Shenzhen}
 \country{China}}
  \email{linkenghong@stu.hit.edu.cn}

\author{Fan Li}
\affiliation{%
  \institution{The Hong Kong Polytechnic University}
  \city{Hong Kong}
  \country{China}}
\email{fan-5.li@polyu.edu.hk}

\author{Pengfei Jia}
\affiliation{%
 \institution{Harbin Institute of Technology}
 \city{Shenzhen}
 \country{China}}
  \email{jiapengfei@stu.hit.edu.cn}
\renewcommand{\shortauthors}{Xinyu Zhang et al.}



\begin{abstract}

Using pre-trained large language models (LLMs) as a backbone for time series prediction has recently attracted growing research interest. Existing approaches typically split time series into patches, map them to the token space of LLMs via a \textbf{Tokenizer}, process the tokens through a frozen or fine-tuned LLM backbone, and then reconstruct numerical forecasts using a \textbf{Detokenizer}. However, the actual effectiveness of LLMs for time series forecasting remains under debate. We observe that when trained and evaluated on small datasets, these Tokenizer–Detokenizer pairs often overfit to the specific data distribution, thereby masking the intrinsic predictive capability of the LLM backbone. To investigate the inherent potential of LLMs in this context, we design three models with identical architectures but distinct pre-training strategies. By leveraging large-scale pre-training, we obtain more unbiased Tokenizer–Detokenizer pairs that are seamlessly integrated with the LLM backbone. Through controlled experiments, we evaluate the zero-shot and few-shot forecasting performance of the LLM, offering insights into its true capabilities. Our extensive experiments reveal that, although the LLM backbone shows some promise, its performance remains limited and does not consistently surpass that of models specifically trained on large-scale time series data. Our source code is publicly available in the repository: https://github.com/SiriZhang45/LLM4TS.

\end{abstract}
\maketitle

\section{Introduction}

Time series forecasting is a long-standing and essential task with applications across numerous fields, including weather forecasting, renewable energy power prediction, stock market forecasting, and disease spread prediction. The evolution of time series forecasting models has progressed from early statistical methods like AutoRegressive Integrated Moving Average (ARIMA)~\cite{box2015time} to advanced deep learning architectures such as RNNs~\cite{salinas2020deepar}, LSTMs, and CNNs~\cite{wang2022micn, wu2022timesnet}, culminating in the latest, powerful Transformer-based models~\cite{nie2022time} and other networks~\cite{zhang2024frnet}. This progression has significantly improved the accuracy and performance of predictive models in various domains. 

In natural language processing, the development of large language models (LLMs) such as GPT~\cite{radford2019language} and LLaMA~\cite{touvron2023llama} has led to recent advancements demonstrating impressive few-shot and zero-shot capabilities in both linguistic and vision-language tasks. Consequently, performing time series prediction tasks based on pre-trained LLMs has emerged as a novel research trend within the time series prediction community~\cite{zhou2023one, tan2024language, hu2025context}. These works focus on adapting pre-trained LLMs to classify, forecast, and detect anomalies in time series data. These studies assume that language models, sophisticated models of sequential dependencies in text, can generalize to sequential dependencies in time series data. Early efforts in leveraging LLMs for time series prediction primarily involved converting numerical values into their character forms and directly feeding them into LLMs to predict future numeric sequences~\cite{xue2023promptcast, gruver2024Large}. More recent advances, however, have shifted toward a more structured paradigm that aligns with conventional time series modeling workflows. In this prevailing approach, time series are divided into patches, which are then tokenized into the token dimensions of LLMs. These tokens are processed through a frozen or fine-tuned LLM backbone, and the final output is detokenized into numerical predictions for future time points~\cite{zhou2023one}. Empirical results on public datasets demonstrate that this paradigm can achieve performance comparable to or even surpassing that of state-of-the-art (SOTA) time series forecasting models. In this paper, we focus on this mainstream paradigm.

\newcommand{\cmark}{\textcolor{green}{\ding{51}}}  
\newcommand{\xmark}{\textcolor{red}{\ding{55}}}

\renewcommand{\arraystretch}{1.6}
\begin{table*}[ht]
\centering
\caption{Comparison of Existing Methods and Our Approach in Time Series Forecasting Using LLMs}
\vspace{-2ex}
\label{tab:method_comparison}
\resizebox{\linewidth}{!}{
\begin{tabular}{l|c|c|c}
\toprule

 & \Large \textbf{Studies Affirming LLM Effectiveness~\cite{zhou2023one,jin2023time,jia2024gpt4mts,liu2024unitime}} & \Large \textbf{Studies Refuting LLM Effectiveness~\cite{tan2024language}} & \Large \textbf{Ours} \\
\midrule
\Large\textbf{Training Data Scale} & \Large Small datasets & \Large Small datasets & \Large \textbf{Large-scale pre-training dataset} \\
\Large\textbf{Testing Setup} & \Large In-distribution, on training-aligned data & \Large In-distribution, on training-aligned data & \Large \textbf{Zero-shot prediction (out-of-distribution)} \\
\Large\textbf{Tokenizer-Detokenizer Coupling} & \Large Strong coupling with backbone & \Large Strong coupling with backbone & \Large \textbf{Decoupled} \\
\Large\textbf{Evaluation of LLM Effectiveness} & \Large \xmark  No quantification & \Large \xmark  No quantification & \Large \cmark \textbf{Quantified} \\

\bottomrule
\end{tabular}
}
\vspace{-2ex}
\end{table*}

Despite the proliferation of methods applying LLMs to time series forecasting, the actual efficacy of LLMs in this domain remains controversial. A recent study~\cite{tan2024language} conducted extensive ablation experiments in which the pre-trained LLM backbone was replaced with simpler alternatives—such as standard Transformer or attention blocks—or even removed entirely. Surprisingly, performance on small datasets remained comparable, and in some cases even improved. The authors attributed these results to architectural factors such as the patching mechanism and channel independence, suggesting that the pre-trained LLM backbone may not be essential for achieving strong forecasting performance. However, these findings do not conclusively demonstrate that LLMs are ineffective. Notably, models incorporating frozen LLMs still achieved or approached SOTA performance on several benchmarks, indicating that pre-trained knowledge may still play a beneficial role. This leaves open the question of whether—and to what extent—LLMs contribute to the observed performance gains. Indeed, several follow-up studies continue to operate under the assumption that LLMs possess inherent capabilities for time series prediction~\cite{wolff2025using, liu2024taming, liu2025efficient, hu2025context}. Therefore, while the work of Tan et al.~\cite{tan2024language} raises important concerns about the necessity of LLMs in time series forecasting, a deeper examination of their experimental design and conclusions is warranted. A more nuanced understanding is needed to disentangle the contributions of model architecture from those of pre-trained representations.

We argue that the common practice in existing studies—training and testing on small datasets using a frozen LLM backbone, with the Tokenizer and Detokenizer trained solely on limited data—is inadequate for assessing the true potential of frozen Transformers in time series forecasting. \emph{Since the Tokenizer and Detokenizer can adapt effectively to the small dataset without relying on pre-trained knowledge, they become highly specialized to the small training data. This tight coupling with the specific dataset masks the intrinsic predictive capability of the LLM backbone, thereby undermining the validity of such evaluations.}

Due to the limitations of conventional train-then-test evaluation on small datasets, zero-shot and few-shot assessment methods offer a more appropriate framework for evaluating LLM capabilities in time series forecasting. However, existing works on such evaluations still suffer from critical flaws~\cite{zhou2023one, tan2024language}. Typically, these methods train the Tokenizer and Detokenizer on a single small dataset with a frozen LLM backbone, then test on a different dataset. When trained on limited data, the Tokenizer-Detokenizer pair learns representations that are highly specific to the source dataset, introducing strong dataset-induced biases. As a result, model performance becomes heavily dependent on the similarity between the training and test data distributions. Conducting zero-shot or few-shot evaluations under such biased conditions leads to misleading assessments of the backbone LLM’s true capabilities, making it difficult to isolate and measure its intrinsic forecasting ability. Empirical evidence supporting these claims is presented in Section~\ref{sec:observation_and_analysis}, including Figures~\ref{f:three_gpt},~\ref{f:under_fitting}, and Table~\ref{tab: encoder preference}.

Given these observations, \emph{we informally define the phenomenon where the Tokenizer-Detokenizer over-adapt to small training datasets, leading to inaccurate evaluations of LLMs' effectiveness in time series forecasting, as Tokenizer-Detokenizer Bias}. To address these limitations, we propose a novel approach based on a consistent model architecture. By keeping the model's architecture unchanged and employing three different pre-training strategies, we aim to obtain more unbiased Tokenizer-Detokenizer pairs through extensive pre-training data exposure. We then conduct zero-shot and few-shot evaluations while carefully controlling the amount of pre-trained knowledge introduced. This enables us to decouple the Tokenizer-Detokenizer from the LLM backbone better and assess the intrinsic capacity of LLMs in time series forecasting tasks. 

A key distinction between our work and prior studies is summarized in Table~\ref{tab:method_comparison}. While previous works either affirm or refute the usefulness of LLMs based on small datasets and in-distribution evaluations, we adopt a novel paradigm that enables zero-shot assessment and quantitative measurement of LLM's effectiveness. The main contributions in this paper are summarized as follows:

\begin{itemize}[leftmargin=1em]
\item We discover that current LLM-based time series prediction models tend to overly adapt to the dataset through their Tokenizer-Detokenizer. This results in a tight coupling between these components and the frozen LLM Backbone when training and testing on small datasets. These limitations hinder the accurate assessment of LLM performance in time series forecasting.
\item To explore the potential of LLMs in time series prediction, we develop three models with identical architectures but distinct pre-training strategies. Pre-training on a large-scale dataset enables us to create unbiased Tokenizer-Detokenizer components tailored to an LLM. Through controlled experiments, we assess the zero-shot and few-shot prediction performance of the LLM, offering insights into its capabilities.
\item  We examine some widely used approaches for improving LLM performance in time series prediction while minimizing the influence of the Tokenizer and Detokenizer. Specifically, we test whether aligning time series tokens with the vocabulary enhances predictions and compare models fine-tuned from LLMs with randomly initialized. Additionally, we quantify the forecasting capabilities of the LLM through experiments.
\end{itemize}

\section{Related Work}

Recently, there has been growing interest in leveraging pre-trained LLMs for time series forecasting, motivated by the hypothesis that LLMs’ rich pre-training on sequential and contextual patterns can enhance modeling of temporal dependencies. A pioneering work, OFA~\cite{zhou2023one}, demonstrated the feasibility of adapting LLMs to time series tasks by partitioning input sequences into patches and embedding them as token-like representations. During training, the Multi-Head Attention and Feed-Forward layers of the LLM backbone are kept frozen, preserving its pre-trained knowledge. Remarkably, OFA achieved performance on par with state-of-the-art time series models across multiple forecasting benchmarks. Subsequent studies extended this framework with more sophisticated fine-tuning strategies, further improving performance. Building upon this paradigm, numerous approaches have introduced architectural and methodological innovations. TimeLLM~\cite{jin2023time} aligns time series patches with LLM word embeddings via cross-attention and enriches inputs with statistical features. TEST~\cite{suntest} employs contrastive learning to bridge the modality gap by aligning time series embeddings with textual prototypes from the LLM's vocabulary. TEMPO~\cite{cao2023tempo} and S2IP-LLM~\cite{pan2024textbf} decompose time series into trend, seasonal, and residual components, using learned word embeddings as semantic anchors for prediction. UniTime~\cite{liu2024unitime} incorporates textual instructions as domain-specific prompts to enable flexible, task-adaptive forecasting. GPT4MTS~\cite{jia2024gpt4mts} fuses numerical time series with auxiliary textual context to improve prediction accuracy. aLLM4TS~\cite{bian2024multi} and LLM4TS~\cite{chang2023llm4ts} adopt a two-stage training strategy: first using autoregressive next-token prediction, followed by fine-tuning for multi-horizon forecasting. CALF~\cite{liu2024taming} proposes a cross-modal knowledge distillation framework with separate branches for time and text modalities, while TimeCMA~\cite{liu2024timecma} integrates dual-modal encoding to generate robust time series representations.

Despite these advances, a recent study by Tan et al.~\cite{tan2024language} challenges the necessity of the pre-trained LLM backbone. Through extensive ablation experiments, they show that replacing the LLM with simpler architectures—such as standard Transformers or attention blocks—does not degrade performance on small datasets; in some cases, it even leads to improvements. These findings raise fundamental questions about the actual contribution of LLMs to forecasting performance, suggesting that architectural design (e.g., patching, channel independence) may play a more critical role than pre-trained knowledge. Nevertheless, this conclusion is not definitive. Several limitations in the experimental setup and evaluation methodology warrant further investigation. Moreover, models incorporating frozen LLMs still achieve or approach state-of-the-art results on certain benchmarks, indicating that pre-trained representations may still offer value. Indeed, many recent works continue to assume the effectiveness of LLMs as a foundational premise~\cite{hu2025context, liu2024taming, liu2025efficient, liu2025efficient, zhao2025enhancing}, reflecting ongoing belief in their potential. For a deeper discussion on the functional role of LLMs in time series forecasting and the validity of current evaluation practices, we refer readers to Appendix~\ref{sec:appendix_2}.



\begin{table*}[ht]
  \centering 
   \caption{Zero-shot Evaluation on ETTH1 of Tokenizer and Detokenizer Trained on Different Datasets}
   \vspace{-2ex}
    \renewcommand{\arraystretch}{1.1}
  \label{tab: encoder preference}
  \resizebox{\textwidth}{!}{
  \begin{tabular}{l|cccccccccc}
    \toprule
    Dataset & ETTH1 & ETTM1 & Exchange rate & Weather &Electricity & 1/10 utsd 1G & 1/2 utsd 1G & utsd 1G & utsd 2G\\
    \midrule
    
    MSE & 0.362 & 0.541 & 1.079 & 0.714 & 0.368 & 0.491 & 0.483 & 0.482 & 0.476\\
    MAE & 0.402 & 0.506 & 0.674 & 0.578 & 0.400 & 0.479 & 0.472 & 0.471 & 0.465\\
    
    \bottomrule
  \end{tabular}}
  \vspace{-2ex}
\end{table*}

\section{Observations and Analysis}
\label{sec:observation_and_analysis}

\subsection{Dataset-Coupling Effects of Tokenizer and Detokenizer}
\label{sec:tokenizer_adaptation}


\begin{figure}[!ht] 
\centering
    \includegraphics[scale=0.34]{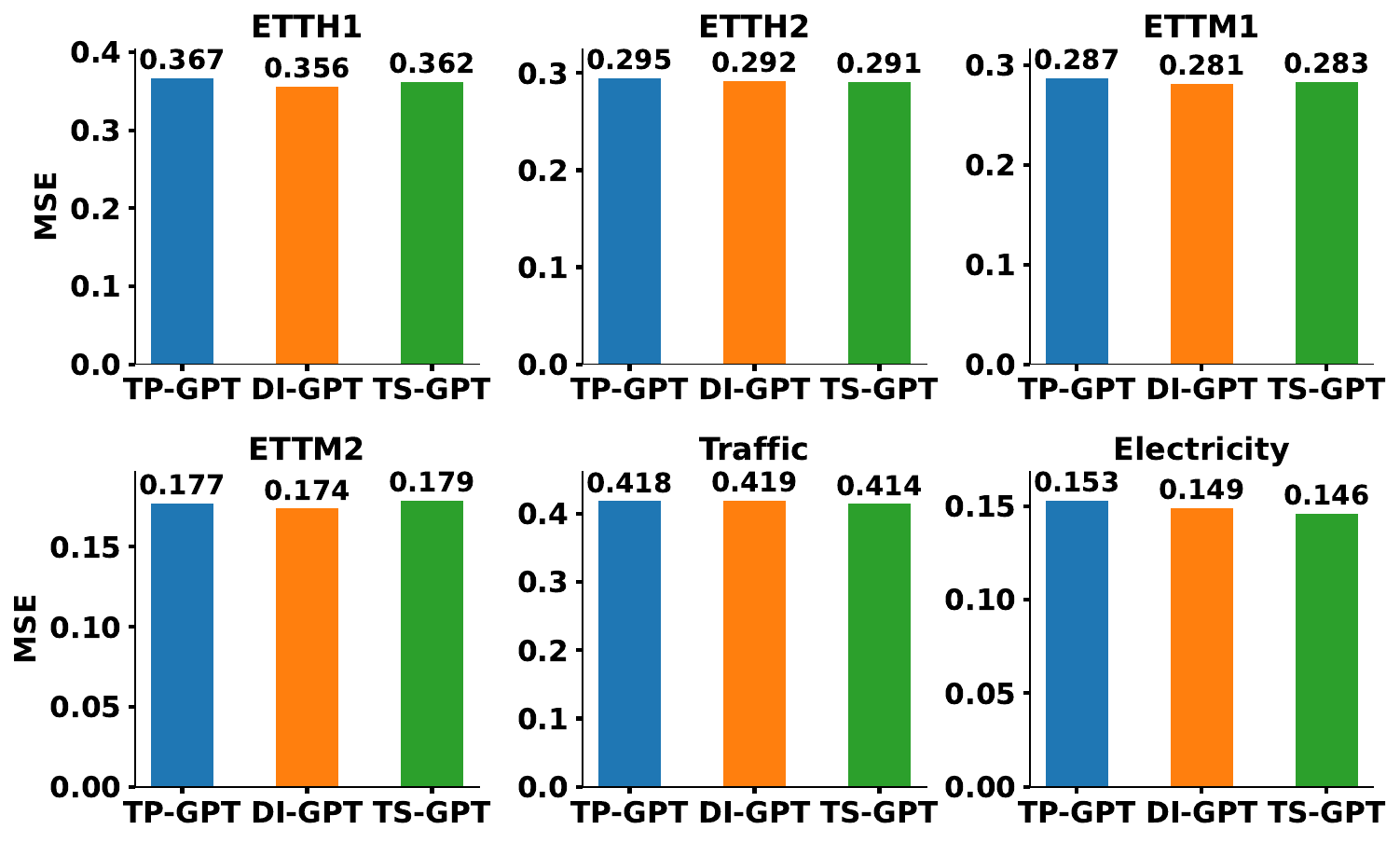}
    \Description{}
    \vspace{-5ex}
    \caption{{Comparison of prediction performance for TP-GPT, DI-GPT, and TS-GPT models on small public datasets. Despite differing pre-training methods, the models exhibit similar performance, indicating that on limited data, the Tokenizer and Detokenizer adapt effectively without relying on additional pre-trained parameters. This suggests the frozen Backbone's potential remains unexplored in such settings.} 
    \label{f:three_gpt}}
\vspace{-2ex}
\end{figure}

To demonstrate the limitations of the conventional train-then-test method on small datasets, we created three comparative models based on a 12-layer GPT2 architecture: The first model, \textit{Textpre-trained GPT (TP-GPT)}, retains the original 12-layer text-pre-trained GPT weights without any changes. The second model, \textit{Diagonal Initialized GPT (DI-GPT)}, initializes all weight matrices of the Transformer layers as diagonal matrices and sets all biases to zero, creating a network with no pre-trained knowledge that mainly passes input data through with minimal non-linear operations. The third model, \textit{Time Series pre-trained GPT (TS-GPT)}, maintains the identical architecture but is randomly initialized and pre-trained on a large time series dataset. This pre-training equips the third model with knowledge specifically beneficial for time series forecasting.

All three models share the same architecture. With their Transformer backbone frozen, we retrained each model's Tokenizer and Detokenizer—both two-layer MLPs—on the training sets of various public datasets and evaluated them on the corresponding test sets. As shown in Figure~\ref{f:three_gpt}, despite differing pre-trained knowledge, the three architecturally identical models showed no significant performance differences during testing. This highlights a critical issue: on small public datasets, \emph{the Tokenizer and Detokenizer can effectively adapt to the data without relying on additional pre-trained parameters}, thus obscuring the frozen LLM backbone's potential. Consequently, this small dataset train-test setup cannot accurately assess LLMs' time series forecasting capabilities. Hence, it suggests that the work~\cite{tan2024language} does not sufficiently challenge the effectiveness of pre-trained LLMs. This is because each model's Patch Tokenizer and Detokenizer are adapted to the current Backbone within the limited dataset context, and the Tokenizer-Detokenizer can provide considerable predictive power alone. As a result, their ablation studies fail to reveal the extent to which the Backbone contributes to processing time series data.

\subsection{Underfitting and Scale Sensitivity of LLM-Based Models}
\label{sec:underfitting_llm}

\begin{figure}[!ht] 
\centering
    \includegraphics[scale=0.4]{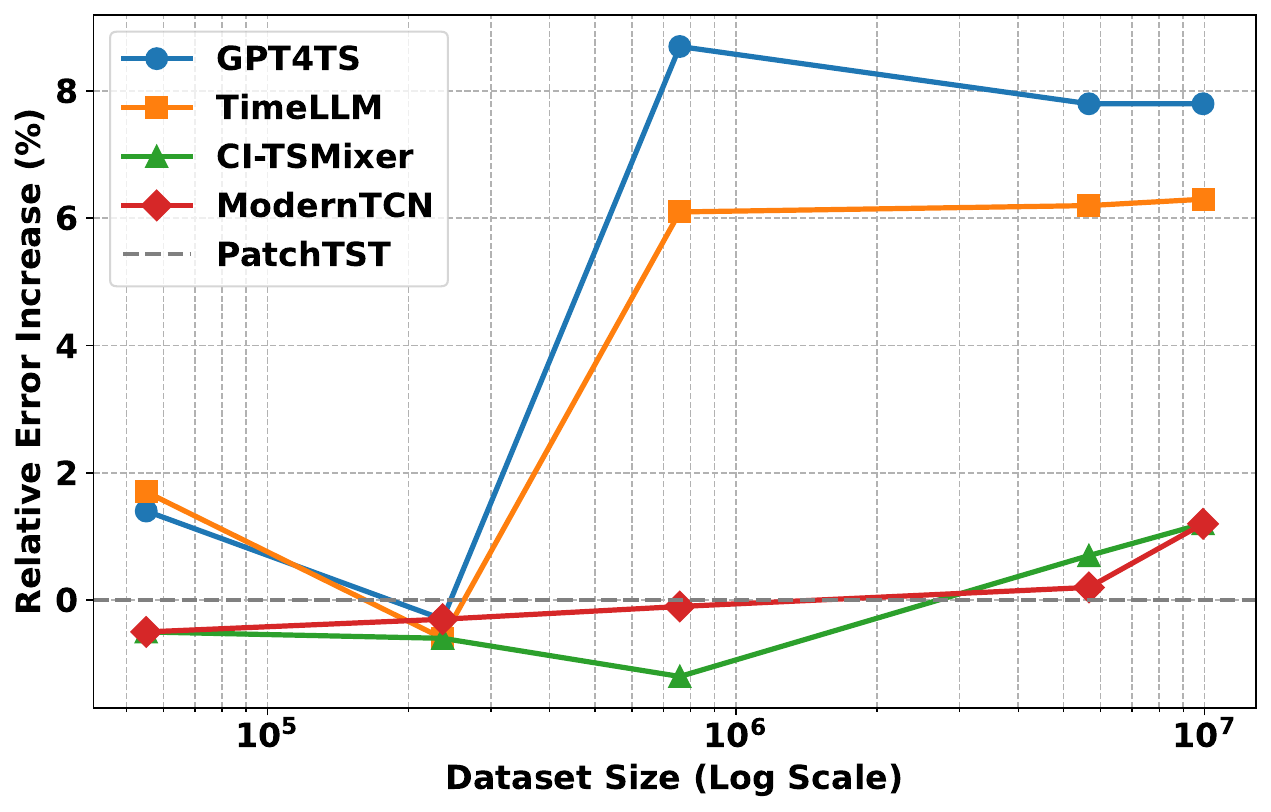}
    \Description{}
    \vspace{-4ex}
    \caption{{LLM-based models exhibit lower relative errors on small datasets but higher relative errors as the dataset size increases. In contrast, TSMixer and ModernTCN maintain consistent performance across all dataset sizes.} 
    \label{f:under_fitting}}
\vspace{-2ex}
\end{figure}

Additionally, we observed that existing LLM-based models commonly suffer from underfitting due to a relatively small number of Tokenizer-Detokenizer parameters. We compared LLM-based models, such as OFA~\cite{zhou2023one} and TimeLLM~\cite{jin2023time}, with the linear model TSMixer~\cite{ekambaram2023tsmixer} and the CNN-based model ModernTCN~\cite{luo2024moderntcn}. These models were trained and tested on public datasets of varying sizes, and their relative errors against PatchTST were measured, as shown in Figure~\ref{f:under_fitting}. For small datasets, OFA's relative error compared to PatchTST is lower. However, as the dataset size increases, the relative error of LLM-based models grows significantly. In contrast, TSMixer and ModernTCN do not exhibit this trend. This implies that in small datasets, existing LLM-based models rely more on the trainable parameters of the Tokenizer and Detokenizer rather than effectively leveraging the pre-trained knowledge of the LLM. These trends are further supported by a theoretical rationale in Appendix~\ref{sec:appendix_3}, which draws from established generalization theories for deep models. Given these findings, we believe zero-shot and few-shot evaluations are essential for decoupling the Tokenizer and Detokenizer from over-adapting to training data and accurately assessing the intrinsic capabilities of LLMs in time series forecasting.

\subsection{Data Biases in Zero-Shot and Few-Shot Evaluations}
\label{sec:zero_shot_bias}

We also argue that existing evaluations of zero-shot and few-shot performance for LLM-based time series prediction have limitations, mainly because the Tokenizer and Detokenizer are not trained in an unbiased manner. Current experiments focus on small, publicly available datasets. For zero-shot testing, researchers typically train the Tokenizer and Detokenizer while freezing the LLM Backbone on one dataset, with some studies also fine-tuning the Backbone. The model is then tested on another small dataset. For few-shot testing, researchers pre-train the Tokenizer and Detokenizer on a small dataset and subsequently fine-tune the Patch Tokenizer and Detokenizer or a subset of parameters using 5\% to 10\% of the target dataset's training data. However, our experiments show that the Patch Tokenizer and Detokenizer trained on small datasets exhibit strong data biases. We froze the Backbone and trained the Tokenizer-Detokenizer on different datasets, then tested their zero-shot performance on the ETTH1 test set (Table~\ref{tab: encoder preference}). The composition of pre-training data significantly impacts performance: better results occur when the training data distribution resembles that of ETTH1. For instance, the Electricity dataset, which shares a similar distribution, yields notably better zero-shot results, while the Exchange Rate dataset, with a very different distribution, performs poorly. Due to the absence of a vocabulary or discrete codebook akin to those in language or image domains~\cite{gu2022vector}, the Tokenizer and Detokenizer fail to learn robust time series representations from small datasets and instead introduce strong biases tied to the training data. As a result, zero-shot and few-shot tests are conducted under significant data representation bias, leading to pre-trained representations that may not generalize to the test data. Consequently, conclusions drawn from such tests lack validity and fail to effectively evaluate the predictive capabilities of LLMs in time series tasks.

In summary, our analysis reveals that current LLM-based forecasting pipelines are hindered by dataset-specific coupling and biased evaluation procedures. In the next section, we introduce a controlled evaluation framework to address these issues and fairly assess the intrinsic capabilities of LLMs.
\section{A Controlled Evaluation Framework}



\begin{figure*}[ht]
\centering
\newlength{\figheight}
\setlength{\figheight}{0.35\textheight} 

\begin{minipage}[b]{0.38\linewidth}
    \centering
    \includegraphics[height=\figheight, keepaspectratio]{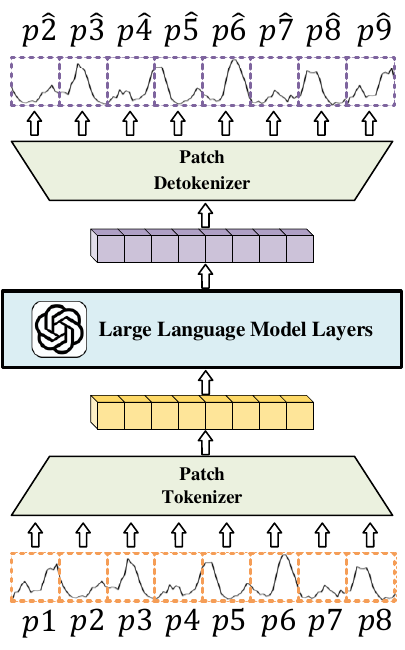} 
    \vspace{-2ex}
    \caption{The identical architecture processes time series patches through tokenization, GPT-2 transformer layers, and detokenization.}
    \label{f: agent model}
\end{minipage}
\hfill
\begin{minipage}[b]{0.61\linewidth}
    \centering
    \includegraphics[height=\figheight, keepaspectratio]{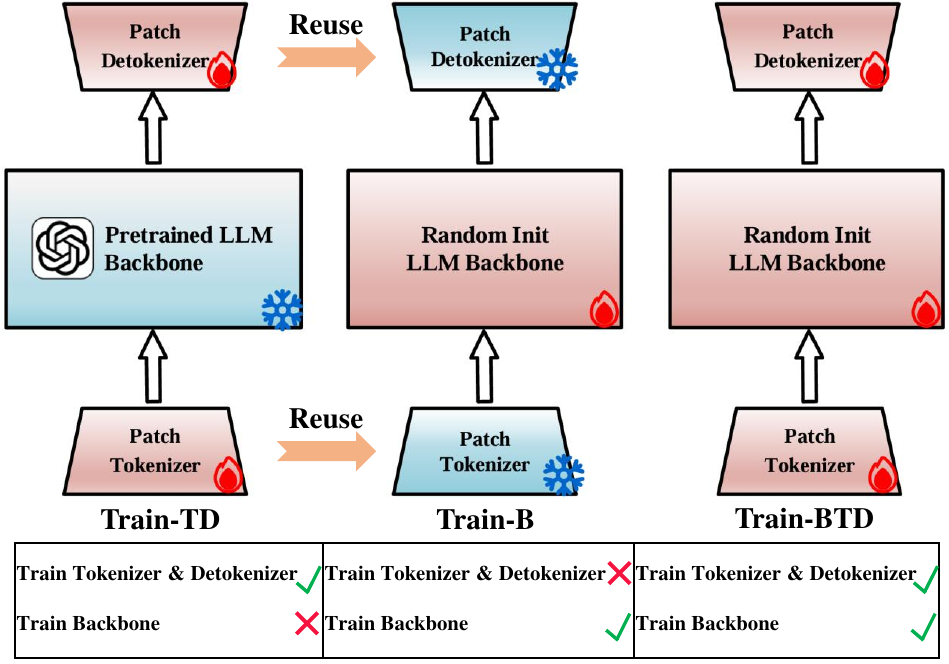} 
    \Description{Three pre-training paradigms: Freeze LLM and train Tokenizer-Detokenizer (Train-TD), reuse Tokenizer-Detokenizer and train backbone (Train-B), and train all parameters (Train-BTD).}
    \vspace{-3ex}
    \caption{Three pre-training paradigms: Freeze LLM and train Tokenizer-Detokenizer (Train-TD), reuse Tokenizer-Detokenizer and train backbone (Train-B), and train all parameters (Train-BTD).}
    \label{f: Framework}
\end{minipage}
\vspace{-2ex}
\end{figure*}

To address the limitations of existing studies, we design three models with identical architectures but different pre-training strategies, enabling a controlled evaluation of LLMs in zero-shot and few-shot time series forecasting. Two main LLM-based architectures are commonly used for time series forecasting: the Encoder-Only~\cite{zhou2023one, jin2023time} and the Decoder-Only~\cite{liu2024autotimes}. We use the autoregressive Decoder-Only architecture since it aligns better with the original generation method of GPT. The identical model architecture is illustrated in Figure~\ref{f: agent model}, where the time series is divided into non-overlapping patches, mapped to the LLM's hidden dimension via a shared tokenizer, resulting in a token sequence. This sequence is processed through backbone layers to yield the hidden states, which is then detokenized to predict subsequent patches in the time series. The model backbone is a 12-layer GPT-2 small-version transformer~\cite{radford2019language}. Based on this architecture, we develop three models using different pre-training strategies. The pre-training models are illustrated in Figure~\ref{f: Framework}.

\textbf{Train-TD (Train Tokenizer-Detokenizer)} employs pre-trained weights from the GPT-2. To reduce biases in the Tokenizer and Detokenizer towards the pre-training dataset, as mentioned earlier, we train the Tokenizer-Detokenizer module using a large pre-training dataset to minimize any potential biases. The backbone layers remain frozen during pre-training, while only the Tokenizer and Detokenizer are trained. This approach preserves the LLM backbone's original predictive capabilities while allowing the trained Tokenizer and Detokenizer to seamlessly adapt the backbone.

\textbf {Train-B (Train Backbone)} reuses the Tokenizer and Detokenizer weights from Train-TD but reinitializes all transformer layer weights. During pre-training, only the transformer layers are trained on the large time series dataset, while the Tokenizer and Detokenizer remain frozen. This aims to replace the text-based pre-training knowledge of GPT-2 with time series' specific knowledge while maintaining consistency with Train-TD's Tokenizer-Detokenizer representation.

\textbf{Train-BTD (Train Backbone and Tokenizer-Detokenizer)} maintains the same architecture as the previous models but initializes all parameters, including the Tokenizer, Detokenizer and Transformer layers. This model is then trained on the large pre-training dataset with all parameters activated. Unlike Train-B—which inherits the Tokenizer-Detokenizer from Train-TD—Train-BTD learns these components jointly with the backbone from scratch, enabling a fully coherent and end-to-end optimized model tailored exclusively to time series data. Given its comprehensive parameter initialization and training approach, this model can be regarded as a large model for time series forecasting. We support the idea that fully pre-trained models on time series data demonstrate strong zero-shot and few-shot predictive capabilities. 

Moreover, the pre-training data excludes the test dataset, facilitating zero-shot and few-shot testing. Additionally, all models are pre-trained on the same data, ensuring consistent pre-training knowledge. Each model employs a two-layer MLP architecture for both the Tokenizer and Detokenizer.
\section{Experiments}
\subsection{Experimental Setup}
\subsubsection{Dataset and baselines.} 
Our pre-training data comes from the UTS dataset, an open-source, large-scale dataset published by \cite{liutimer}. This dataset includes 1G, 2G, 4G, and 12G versions, featuring high-quality time series data from various domains such as energy, IoT, nature, networks, and health. Each version contains challenging time series samples that help the model learn complex patterns of time dependencies. Our research primarily uses the 2G version, which comprises approximately 100 million training samples. This dataset will be used to train: the unbiased Tokenizer-Detokenizer of Train-TD, the Backbone of Train-B, and the full parameters of Train-BTD and other large model baselines. We conduct zero-shot and few-shot experiments on seven real-world datasets for evaluation. These datasets are all multivariate time series from various fields, with their descriptions and statistics summarized in Appendix~\ref{sec:appendix_1}.

In the zero-shot prediction experiments, in addition to our three pre-training models, we fully trained three open-source large time series models—Timer~\cite{liutimer}, Moirai~\cite{woo2024unified}, and Chronos~\cite{ansari2024chronos}—using the same pre-training data as baselines. Timer adopts a Decoder-Only architecture, Moirai utilizes an Encoder-Only architecture, and Chronos employs an Encoder-Decoder architecture. Minor modifications were made to Moirai to accommodate univariate datasets, ensuring compatibility with our experimental setup. To ensure a fair comparison, we select the smallest version of each model, configured with the same number of layers and hidden dimensions as GPT-2, resulting in comparable parameter counts across models. We also include Diagonal Initialized GPT (DI-GPT) as another baseline. DI-GPT initializes all weight matrices of the 12-layer Transformer to diagonal matrices and sets all biases to zero, then freezes the backbone parameters. DI-GPT essentially passes input data directly to the output with minimal nonlinear transformations, serving as a baseline without pre-trained knowledge. Only the Tokenizer-Detokenizer of DI-GPT is trained, ensuring that it has unbiased Tokenizer-Detokenizer components without any pre-training knowledge in the backbone. To fairly compare performance, the Mean Squared Error (MSE) and Mean Absolute Error (MAE) are adopted as evaluation metrics for all experiments.

\subsubsection{Implementation details.} 
Each pre-training model undergoes both zero-shot and few-shot prediction experiments. In the zero-shot experiment phase, we directly test the pre-trained model on the test set of each small dataset without any fine-tuning. In the few-shot experiment phase, we utilize 10\% of the training data from each dataset and conduct testing on the test set. In the fine-tuning experiment, there are two points worth noting: (1) we conducted two separate experiments---one where we froze the Backbone while fine-tuning only the Tokenizer-Detokenizer, and another where we froze the Tokenizer-Detokenizer while fine-tuning only the Backbone; and (2) during fine-tuning, we utilized grid search to select the optimal learning rate for each model, trained for as many epochs as possible with an early stopping criterion set to 3, to maximize the fine-tuning potential of each model on every dataset. To ensure a fair comparison, we set the look-back window $ L $ to 672 and fixed the forecasting horizon $H$ at 96 for all models. To reduce training overhead, the patch size was consistently set to 96. Our experiments present results only for predicting 96 steps, with each experiment repeated five times and the results averaged. In the comparison of the three pre-training models, the best-performing results are emphasized in bold. Additionally, we also evaluated the impact of varying patch sizes and forecast horizons. Detailed results are provided in Appendix~\ref{sec:appendix_6}. While some configurations yield minor differences, our main findings remain consistent.

\subsection{Main Results.}

\begin{table*}
  \vspace{-1ex}
  \caption{Zero-shot Evaluation of Three Models and Baselines}
  \vspace{-2ex}
  \centering
  \resizebox{\textwidth}{!}{
  \renewcommand{\arraystretch}{1.3}
  \begin{tabular}{lcccccc|cccccccc}
    \toprule
     \multirow{2}{*}{\textbf{Dataset}} 
     & \multicolumn{2}{c}{Train-TD} & \multicolumn{2}{c}{Train-B} & \multicolumn{2}{c}{Train-BTD} & \multicolumn{2}{c}{Timer} & \multicolumn{2}{c}{Moirai} & \multicolumn{2}{c}{Chronos} & \multicolumn{2}{c}{DI-GPT}\\
    \cmidrule(lr){2-3}\cmidrule(lr){4-5}\cmidrule(lr){6-7}\cmidrule(lr){8-9}\cmidrule(lr){10-11}\cmidrule(lr){12-13}\cmidrule(lr){14-15}

    & MSE & MAE & MSE & MAE & MSE & MAE & MSE & MAE & MSE & MAE & MSE & MAE & MSE & MAE\\  
    
    \midrule
    \textbf{ETTH1} & 0.476 & 0.465 & 0.427 & 0.432 & \textbf{0.417} & \textbf{0.428} & 0.412 & 0.426 & 0.416 & 0.423 & 0.426 & 0.430 & 0.802 & 0.613\\
    
    \textbf{ETTH2} & 0.273 & 0.339 & 0.291 & 0.347 & \textbf{0.269} & \textbf{0.332} & 0.288 & 0.340 & 0.290 & 0.344 & 0.289 & 0.345 & 0.381 & 0.421\\
    
    \textbf{ETTM1} & 0.542 & 0.473 & \textbf{0.419} & \textbf{0.415} & 0.434 & 0.424 & 0.438 & 0.426 & 0.409 & 0.402 & 0.449 & 0.439 & 0.745 & 0.569\\
    
    \textbf{ETTM2} & 0.209 & 0.292 & 0.220 & 0.296 & \textbf{0.199} & \textbf{0.281} & 0.206 & 0.284 & 0.216 & 0.288 & 0.215 & 0.287 & 0.323 & 0.375\\

    \textbf{Traffic} & 0.639 & 0.429 & 0.577 & 0.400 & \textbf{0.508} & \textbf{0.376} & 0.490 & 0.364 & 0.488 & 0.358 & 0.507 & 0.380 & 1.508 & 0.842\\
    
    \textbf{Electricity} & 0.263 & 0.360 & 0.232 & 0.336 & \textbf{0.198} & \textbf{0.300} & 0.211 & 0.309 & 0.189 & 0.293 & 0.203 & 0.304 & 0.921 & 0.781\\
    
    \textbf{Exchange rate} & 0.131 & 0.249 & 0.128 & 0.251 & \textbf{0.119} & \textbf{0.239} & 0.114 & 0.233 & 0.109 & 0.227 & 0.121 & 0.244 & 0.553 & 0.564\\
    
    \bottomrule
    \label{tab:zeroShotResults}
  \end{tabular}}
  \vspace{-4ex}
\end{table*}
\newcommand{\rowfont}[1]{#1}

\begin{table*}
\vspace{-1ex}
  \caption{Few-shot Evaluation of Three Models with Different Fine-tuning Strategies}
  \vspace{-2ex}
  \centering
  \resizebox{\textwidth}{!}{
  \renewcommand{\arraystretch}{1.5} 
  
  \begin{tabular}{lcccccccccccc}
    \toprule
     \multirow{4}{*}{\Large \textbf{Dataset}} 
     & \multicolumn{4}{c}{\large Train-TD} & \multicolumn{4}{c}{\large Train-B} & \multicolumn{4}{c}{\large Train-BTD}\\
     
\cmidrule(lr){2-5}\cmidrule(lr){6-9}\cmidrule(lr){10-13}

& \multicolumn{2}{c}{\makecell{Fine-tune \\ Backbone}}
& \multicolumn{2}{c}{\makecell{Fine-tune \\ Token-Detoken}}
& \multicolumn{2}{c}{\makecell{Fine-tune \\ Backbone}}
& \multicolumn{2}{c}{\makecell{Fine-tune \\ Token-Detoken}}
& \multicolumn{2}{c}{\makecell{Fine-tune \\ Backbone}}
& \multicolumn{2}{c}{\makecell{Fine-tune \\ Token-Detoken}} \\

\cmidrule(lr){2-3}\cmidrule(lr){4-5}\cmidrule(lr){6-7}\cmidrule(lr){8-9}\cmidrule(lr){10-11}\cmidrule(lr){12-13}

    & \large MSE & \large MAE & \large MSE & \large MAE & \large MSE & \large MAE & \large MSE & \large MAE & \large MSE & \large MAE & \large MSE & \large MAE \\
    
\midrule
\large \textbf{ETTH1} & \large 0.435 & \large 0.440 & \large 0.455 & \large 0.453 & \large 0.408 & \large 0.422 & \large 0.423 & \large 0.428 & \large 0.414 & \large 0.427 & \large \textbf{0.405} & \large \textbf{0.420} \\
\large \textbf{ETTH2} & \large 0.264 & \large 0.332 & \large 0.270 & \large 0.338 & \large 0.271 & \large 0.332 & \large 0.269 & \large 0.335 & \large 0.261 & \large 0.329 & \large \textbf{0.258} & \large \textbf{0.327} \\
\large \textbf{ETTM1} & \large 0.428 & \large 0.414 & \large 0.512 & \large 0.451 & \large 0.392 & \large 0.406 & \large \textbf{0.388} & \large \textbf{0.401} & \large 0.409 & \large 0.417 & \large 0.399 & \large 0.414 \\
\large \textbf{ETTM2} & \large 0.205 & \large 0.289 & \large 0.198 & \large 0.279 & \large 0.196 & \large 0.278 & \large 0.196 & \large 0.271 & \large 0.189 & \large 0.268 & \large \textbf{0.185} & \large \textbf{0.265} \\
\large \textbf{Traffic} & \large 0.509 & \large 0.380 & \large 0.492 & \large 0.346 & \large 0.512 & \large 0.384 & \large 0.461 & \large 0.334 & \large 0.471 & \large 0.358 & \large \textbf{0.457} & \large \textbf{0.329} \\
\large \textbf{Electricity} & \large 0.178 & \large 0.280 & \large 0.180 & \large 0.280 & \large 0.176 & \large 0.280 & \large 0.168 & \large 0.269 & \large 0.154 & \large 0.258 & \large \textbf{0.154} & \large \textbf{0.257} \\
\large \textbf{Exchange rate} & \large 0.109 & \large 0.228 & \large 0.120 & \large 0.237 & \large 0.118 & \large 0.236 & \large 0.098 & \large 0.217 & \large 0.090 & \large 0.217 & \large \textbf{0.089} & \large \textbf{0.209} \\
    
    \bottomrule
  \end{tabular}
  }
  \label{tab:fewShotResultsSelected}
  \vspace{-2ex}
\end{table*}

The zero-shot test results for the three models and multiple baselines are shown in Table~\ref{tab:zeroShotResults}. The DI-GPT performs significantly worse than all other pre-training models in zero-shot prediction. This indicates that, without any pre-training knowledge, relying solely on the Tokenizer-Detokenizer is insufficient for effective zero-shot predictions. This result supports our motivation for comparing models: conducting zero-shot testing with sufficient pre-training can prevent the Tokenizer-Detokenizer from becoming adapted to the specific dataset without relying on the backbone parameter. The superior zero-shot prediction ability of Train-TD compared to DI-GPT also suggests that LLM-based models possess a certain degree of capability in modeling time series data. Despite having the same architecture, the three pre-training models exhibited different performance levels: Train-BTD achieved the best results, followed by Train-B, while Train-TD performed the worst. The performance differences were particularly pronounced on large datasets such as Traffic and Electricity. These results indicate that time series prediction models based on language models are less capable than large time series models trained on 100 million samples.

Comparing Train-TD and Train-B, we observe that replacing the frozen pre-trained text knowledge with time series knowledge significantly improves prediction performance on most datasets. This suggests that pre-trained text knowledge is not particularly effective for time series prediction tasks. However, this improvement is not observed in the ETTH2 and ETTM2 datasets containing significant noise and distribution shifts. This is mainly because models that focus more on short-term temporal dependencies tend to perform better on noisier datasets compared to those capable of modeling longer-term dependencies. Conversely, this phenomenon highlights the weakness of time series forecasting models based on language models in handling complex long-term dependencies. We provide detailed experimental and theoretical evidence for this observation in Appendix~\ref{sec:appendix_4}. Comparing Train-B and Train-BTD, we observe that Train-BTD significantly outperforms Train-B on most datasets, despite sharing the same pre-trained knowledge. This indicates that the Tokenizer-Detokenizer adapted for LLM backbones is not well-suited for time series prediction tasks. Comparing the zero-shot test results of Train-BTD, Chronos, Timer, and Moirai, we find that the Encoder-Only and Decoder-Only architectures each have distinct advantages across different datasets. All four models can be considered viable foundation models for time series analysis. Overall, the Encoder-Only architecture demonstrates superior performance on more complex and high-dimensional datasets, such as the Traffic and Weather datasets. This suggests that the bidirectional context modeling inherent in encoder-only models is particularly effective in capturing rich temporal structures in challenging forecasting scenarios. In contrast, the Chronos, which adopts a discrete binning strategy for numerical value representation and employs an Encoder-Decoder architecture primarily designed for sequence-to-sequence generation, exhibits slightly inferior performance compared to the Encoder-Only and Decoder-Only counterparts. This performance gap may stem from the information loss introduced by quantization in the binning process, as well as a potential mismatch between the generative pre-training objective and the zero-shot forecasting task, which aligns with findings from existing evaluations~\cite{yao2024towards}. A detailed comparison of these models is beyond the scope of our current work. For external validation of our findings, we evaluated the zero-shot capabilities of Train-TD and Train-B, and Train-BTD on a larger public benchmark — the open leaderboard proposed by Ansari et al. (2024), as detailed in Appendix~\ref{sec:appendix_7}. And the results align well with the above conclusions.

The few-shot test results for the three pre-training models are shown in Table~\ref{tab:fewShotResultsSelected}. Overall, fine-tuning the Tokenizer-Detokenizer of Train-BTD achieves the best prediction performance. Models based on time series knowledge demonstrate greater fine-tuning potential than Train-TD, which shows limited improvement regardless of the fine-tuning method. 
When comparing the fine-tuning of the Tokenizer-Detokenizer with the Transformer Backbone, we observe that for Train-TD, fine-tuning the Backbone generally produces better results across most datasets than fine-tuning only the Tokenizer-Detokenizer. Conversely, Models B and C perform better on most datasets when only the Tokenizer-Detokenizer is fine-tuned. These findings suggest that Train-TD's bottleneck lies in its Backbone pre-training knowledge, whereas Train-BTD's Backbone already possesses sufficient time series knowledge. For Train-BTD, the key factor is adjusting the representation for each dataset to leverage the Backbone better. 

\subsection{Detailed Experimental Analysis}

To provide comprehensive analyses of LLMs for time series forecasting, we further address the following four questions.

\textbf{\textit{Question 1: Can LLM Effectively Leverage Its Vocabulary for Time Series Prediction?}}

\begin{figure}[htbp]  
    \vspace{-2ex}
    \centering  
    \Description{}
    \subfigure[]{  
        \begin{minipage}[b]{0.49\linewidth}  
            \centering  
            \includegraphics[width=\linewidth]{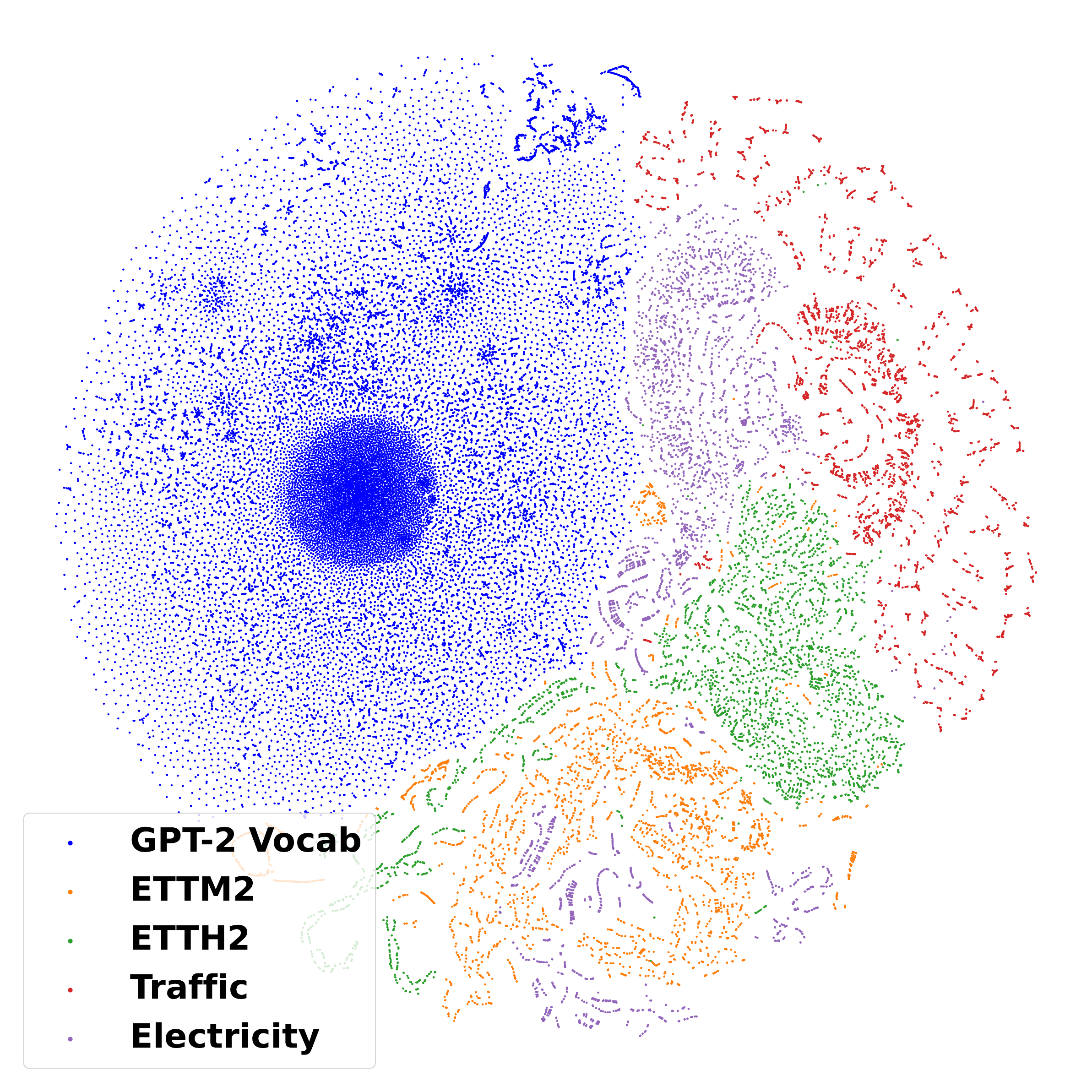}  
            \vspace{-3ex} 
            \label{f:gpt vocab}  
        \end{minipage}  
    }%
    \hspace{-0.7em}
    \subfigure[]{  
        \begin{minipage}[b]{0.49\linewidth}  
            \centering  
            \includegraphics[width=\linewidth]{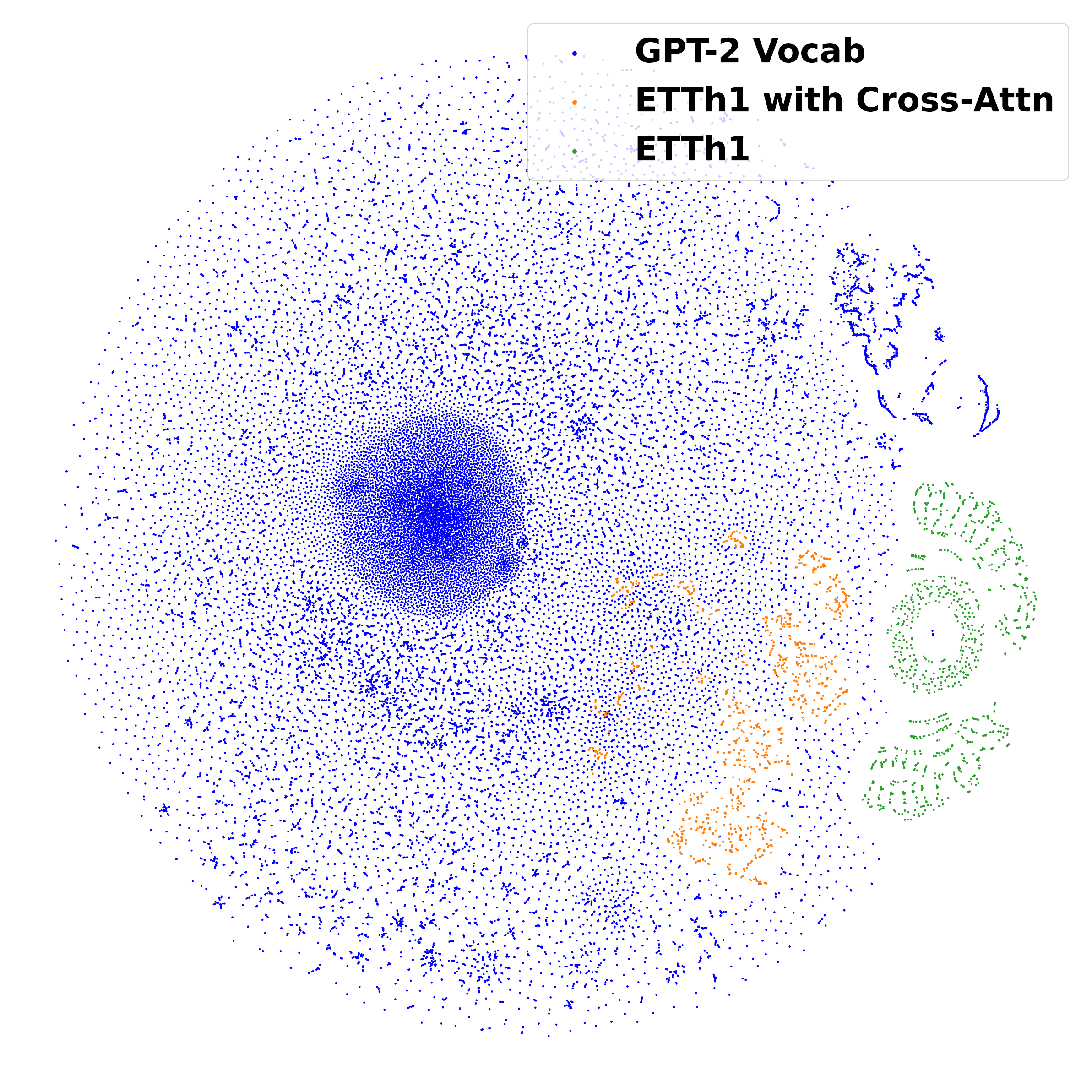}  
            \vspace{-3ex} 
            \label{f:cross attn}  
        \end{minipage}  
    }

    \vspace{-3ex}
    \caption{(a) Illustrates the distribution of GPT-2's vocabulary tokens alongside token representations from various time series datasets, showing a clear separation between these two distributions. (b) Demonstrates tokens from the ETTH1 dataset are mapped from outside to inside the GPT-2 vocabulary range using a cross-attention mechanism.}    
    \vspace{-3ex}
\end{figure}

LLMs operate by modeling relationships within their vocabulary, implying that time series tokens need to be within the area of the model's vocabulary for the LLM pre-trained backbone to be effectively leveraged. We applied t-SNE dimensionality reduction to the token representations from Train-TD's Tokenizer for each dataset and compared them with the token representations from GPT-2's vocabulary, illustrated in Figure~\ref{f:gpt vocab}. The results show that the Tokenizer-Detokenizer adapted for the LLM maps time series patches into areas outside the distribution of the vocabulary, with a clear boundary and nearly no overlap. This indicates that when Train-TD performs time series prediction tasks, it relies on out-of-distribution generalization rather than leveraging its language modeling capabilities.


Several studies, such as TimeLLM~\cite{jin2023time} and TEST~\cite{suntest}, have attempted to match LLM vocabulary with time series tokens, aiming to align time series data with the vocabulary tokens. However, these methods are often tested on small datasets, which can result in the Tokenizer-Detokenizer being forcibly adapted to specific vocabularies chosen for those datasets. This is similar to how the Tokenizer-Detokenizer may fit the frozen LLM backbone, making it difficult to validate whether a universal vocabulary can enhance the capabilities of LLM-based time series prediction models through such tests. We added TimeLLM's cross-attention module~\cite{jin2023time} to Train-TD's Tokenizer, mapping time series tokens to a set of 1,000 vocabulary items, and then pre-trained the model using the same pre-training data while keeping the Backbone frozen. The results of the zero-shot prediction, shown in the left half of Table~\ref{tab:cross-vocal}, along with the token representations after cross-attention, illustrated in Figure~\ref{f:cross attn}, demonstrate that while time series tokens were mapped into the GPT-2 vocabulary space, the zero-shot prediction performance significantly decreased. We also evaluated an alternative token alignment strategy commonly used in prior work; detailed results are provided in Appendix~\ref{sec:appendix_8}. Across both approaches, the findings are consistent: LLMs employ a form of out-of-domain generalization for time series prediction, which fundamentally differs from their approach to vocabulary modeling. Knowledge acquired through vocabulary modeling is difficult to transfer to time series prediction tasks effectively.



\begin{table}[htbp]
  \caption{Comparing Train-TD, Train-B, Cross-attention Enhanced Train-TD, and Fine-tuned LLM Backbone under Zero-shot Evaluation (MSE Metrics)}
  \vspace{-2ex}
  \label{tab:cross-vocal}
  \scalebox{0.84}{
  \begin{tabular}{lcc|cc}
    \toprule
    \multirow{1}{*}{\makecell{\textbf{Dataset}}} &
    \multicolumn{1}{c}{\makecell{\textbf{Train-TD}}} &
    \multicolumn{1}{c}{\makecell{\textbf{Train-TD with}  \\ \textbf{Cross-Attn}}} & \multicolumn{1}{c}{\makecell{\textbf{Train-B}}} & 
    \multicolumn{1}{c}{\makecell{\textbf{Pretrained} \\ \textbf{from LLM}}} \\

    \midrule
    \textbf{ETTH1} & \LARGE 0.476 & \LARGE 0.782 & \LARGE 0.427 & \LARGE 0.440\\
    \textbf{ETTH2} & \LARGE 0.273 & \LARGE 0.373 & \LARGE 0.291 & \LARGE 0.290\\
    \textbf{ETTM1} & \LARGE 0.542 & \LARGE 0.731 & \LARGE 0.419 & \LARGE 0.418\\
    \textbf{ETTM2} & \LARGE 0.209 & \LARGE 0.318 & \LARGE 0.220 & \LARGE 0.222\\
    \textbf{Traffic} & \LARGE 0.639 & \LARGE 1.465 & \LARGE 0.577 & \LARGE 0.584\\
    \textbf{Electricity} & \LARGE 0.263 & \LARGE 0.896 & \LARGE 0.233 & \LARGE 0.236\\
    \textbf{Exchange rate} & \LARGE 0.131 & \LARGE 0.540 & \LARGE 0.128 & \LARGE 0.129\\
    \bottomrule
  \end{tabular}
  }
  \vspace{-1ex}
\end{table}

\textbf{\textit{Question 2: Does Fine-tuning an LLM Backbone Provide Additional Performance Improvements?}}

Starting with Train-TD, we froze the Tokenizer-Detokenizer and fine-tuned its Backbone using the same data used to train Train-B. As shown in the right half of Table~\ref{tab:cross-vocal}, the model fine-tuned from the LLM backbone does not demonstrate a significant performance advantage over Train-B, a randomly initialized and trained backbone. The performance metrics for both models are quite similar. These results suggest that, in this context, fine-tuning from an LLM backbone may offer little additional benefits compared to training from a randomly initialized backbone. 

\textbf{\textit{Question 3: How Many Time Series Samples Are Needed for Pre-training to Match GPT-2’s Prediction Performance?}}
\begin{figure}[htbp] 
\vspace{-1ex}
\centering
    \includegraphics[scale=0.48]{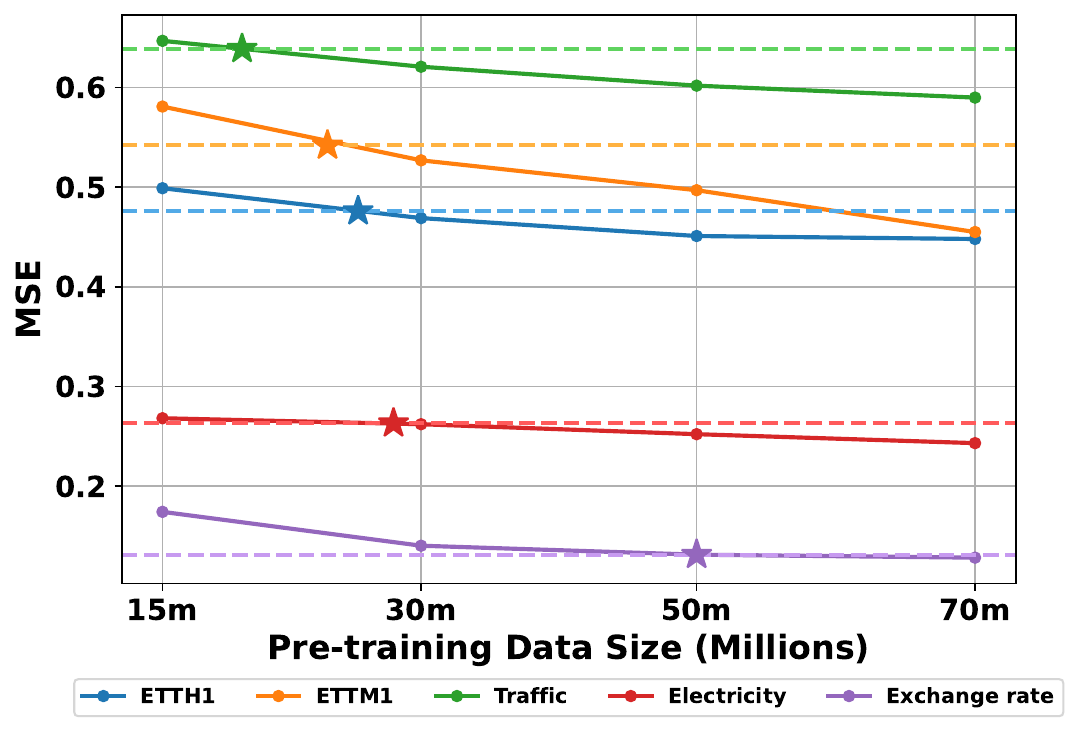}
    \Description{}
    \vspace{-4ex}
    \caption{{Quantification of GPT-2 Backbone's Equivalent Time Series Pre-training Samples} 
    \label{f:quantify gpt}}
\vspace{-2ex}
\end{figure}

We initialized the Transformer of Train-B and trained it using varying amounts of pre-training data, followed by zero-shot prediction experiments. We aimed to measure how much time series pre-training data is needed for Train-B to match Train-TD's predictive performance under the same Tokenizer-Detokenizer conditions. As shown in the figure~\ref{f:quantify gpt}, zero-shot prediction performance improves with more pre-training samples, following the scaling law. The intersection points between Train-B's performance curve and Train-TD's results indicate the required sample size for equivalent performance. This point varies slightly across datasets but generally falls between \emph{15 and 50 million} samples, a relatively small amount. These findings suggest that for tasks involving time series prediction, a Transformer backbone trained from scratch on approximately 50 million time-series samples can achieve predictive performance comparable to that of a frozen GPT-2 backbone. Given its large parameter size, this indicates that LLM's effectiveness for time series forecasting is not high. While this does not necessarily imply that GPT-2's backbone has reached its limit, it does highlight that significant performance improvements may require either substantially more time series data or additional optimizations specifically designed for time series forecasting. 

\textbf{\textit{Question 4: Will a more powerful language model lead to better prediction performance?}} \label{sec:appendix_5}

To evaluate the generalizability of our findings across diverse LLM scales, we conduct a comprehensive comparison using three representative models: GPT-2 (128M), Qwen-1.8B, and LLaMA3-8B. Qwen-1.8B, and LLaMA3-8B following the model configuration outlined in TimeLLM~\cite{jin2023time}. All models are integrated into the same forecasting framework following the \texttt{Train-TD} pre-training strategy, where only the tokenizer and detokenizer are trained on time series data while the backbone LLM remains frozen.  The results of zero-shot forecasting on seven real-world time series datasets are summarized in Table~\ref{tab: GPT_Qwen_LLaMA3}. As shown, when evaluated using both MSE and MAE metrics, neither the Qwen-1.8B nor the LLaMA3-8B model outperforms the GPT-2 counterpart; in most cases, their performances are slightly worse. These results collectively suggest that stronger pre-trained language models—despite their enhanced reasoning and language understanding abilities—do not necessarily generalize better to zero-shot time series forecasting. We hypothesize that this stems from a fundamental misalignment between textual and temporal semantics: the inductive biases learned during language modeling may not effectively transfer to capturing the predictive patterns in time series data.

\begin{table}[h]
  \centering
  \vspace{-1ex}
  \caption{Zero-shot Evaluation across Different LLM Backbones}
  \vspace{-2ex}
  \scalebox{0.92}{
  \begin{tabular}{lcccccc}
    \toprule
     \multirow{2}{*}{\textbf{Dataset}} 
     & \multicolumn{2}{c}{GPT-2-128M} 
     & \multicolumn{2}{c}{Qwen-1.8B} 
     & \multicolumn{2}{c}{LLaMA3-8B} \\
    \cmidrule(lr){2-3}\cmidrule(lr){4-5}\cmidrule(lr){6-7}
    & MSE & MAE & MSE & MAE & MSE & MAE \\  
    \midrule
    \textbf{ETTH1} & \textbf{0.476} & \textbf{0.465} & 0.483 & 0.472 & 0.491 & 0.483 \\
    \midrule
    \textbf{ETTH2} & \textbf{0.273} & \textbf{0.339} & 0.281 & 0.351 & 0.289 & 0.362 \\
    \midrule
    \textbf{ETTM1} & 0.542 & 0.473 & 0.528 & 0.458 & \textbf{0.501} & \textbf{0.435} \\
    \midrule
    \textbf{ETTM2} & \textbf{0.209} & \textbf{0.292} & 0.215 & 0.301 & 0.220 & 0.306 \\
    \midrule
    \textbf{Traffic} & 0.639 & 0.429 & 0.625 & 0.418 & \textbf{0.617} & \textbf{0.407} \\
    \midrule
    \textbf{Electricity} & \textbf{0.263} & \textbf{0.360} & 0.268 & 0.366 & 0.273 & 0.371 \\
    \midrule
    \textbf{Exchange rate} & \textbf{0.131} & \textbf{0.249} & 0.133 & 0.251 & 0.132 & 0.253 \\
    \bottomrule
  \end{tabular}
  }
  \vspace{-3ex}
  \label{tab: GPT_Qwen_LLaMA3}
\end{table}

\section{Conclusion}

In this work, we find that existing time series prediction models based on LLMs often couple the Tokenizer and Detokenizer with a frozen Backbone, especially when trained and tested on small public datasets. This coupling limits the effective evaluation of the LLM's forecasting capability. To address this issue, we introduce three pre-training models with identical architectures but different pre-training strategies. By using large-scale pre-training, we decouple the Tokenizer-Detokenizer from dataset-specific biases, allowing for more controlled assessments. Extensive experiments assessing the zero-shot and few-shot forecasting performance of the LLM backbone demonstrate that its potential is inherently limited.


\bibliographystyle{ACM-Reference-Format}
\balance
\bibliography{ref}
\newpage
\appendix
\appendix

\section{Dataset Details.}\label{sec:appendix_1}
We conducted extensive experiments on seven real-world datasets to evaluate our approach. These datasets encompass multivariate time series from various domains, with their statistics summarized in Table~\ref{tab:statistics}. (1) {\textbf{ETT~\cite{zhou2021informer}}} comprises two sub-datasets, ETTH and ETTM, which record the temperature changes of electric transformers from July 2016 to July 2018. The data are collected at frequencies of one hour and 15 minutes, respectively. (2) {\textbf{Traffic~\cite{lai2018modeling}}} describes the hourly road occupancy rates measured by sensors located on freeways in the San Francisco Bay Area. (3) {\textbf{Electricity~\cite{lai2018modeling}}} contains hourly electricity consumption records for 321 customers, spanning from 2012 to 2014. (4) {\textbf{Exchange rate~\cite{zhou2021informer}}} includes daily exchange rate data for eight countries: Australia, the United Kingdom, Canada, Switzerland, China, Japan, New Zealand, and Singapore, covering the period from 1990 to 2016. All multivariate datasets were fine-tuned and tested using a channel-independent approach~\cite{nie2022time}.

\begin{table}[!ht]
  \centering 
  \caption{Statistics of Datasets in Our Experiments}
  \label{tab:statistics}
  \begin{tabular}{l|ccc}
    \toprule
    Dataset & Sequence Length & Features & Frequency \\
    \midrule
    ETTH1 & 17,420 & 7 & 1h \\
    ETTH2 & 17,420 & 7 & 1h \\
    ETTM1 & 69,680 & 7 & 15 min \\
    ETTM2 & 69,680 & 7 & 15 min \\
    Traffic & 17,544 & 862 & 1h \\
    Electricity & 26,304 & 321 & 1h \\
    Exchange rate & 7588 & 8 & 1 day \\
    \bottomrule
  \end{tabular}
\end{table}

\section{Discussion on the Functionality of LLMs for Time Series Forecasting.}\label{sec:appendix_2}

In the study by Zhou et al. \cite{zhou2023one}, the authors propose that the backbone of pre-trained LLMs can serve as a robust compression module, with the self-attention mechanisms exhibiting behavior akin to Principal Component Analysis (PCA). They argue that the Transformer architecture of LLMs provides a versatile tool for analyzing nearly any sequential data. Specifically, the compression achieved through LLM Transformers enhances the signal-to-noise ratio in time series representations. Their paper informally demonstrates that Transformer models leverage attention mechanisms to perform complex n-gram estimations and that these n-gram distributions' generation rules are broadly applicable. This generality, according to the authors, enables LLMs to achieve cross-domain knowledge transfer in time series prediction tasks.

Mirchandani et al. \cite{mirchandani2023large} further explore the potential of LLMs as universal pattern recognizers through a series of experiments across three domains: sequence transformation, sequence completion, and sequence improvement. Their findings indicate that LLMs can generalize certain sequence transformations effectively. Notably, pre-trained LLMs can autoregressively complete complex token sequences, ranging from arbitrary sequences generated by probabilistic context-free grammar (PCFG) programs to richer spatial patterns found in the Abstract Reasoning Corpus (ARC). These results suggest that LLMs, without additional training, can function as general sequence modelers driven by contextual learning.

However, Tan et al. \cite{tan2024language} challenge the efficacy of using pre-trained LLMs for time series prediction under the current paradigm. After evaluating three popular LLM-based time series prediction methods, they discovered that removing or replacing the LLM component with basic attention layers did not degrade performance; in many cases, it even improved outcomes. Despite the vast number of parameters and high computational costs associated with LLM-based models, pre-trained LLMs did not outperform models trained from scratch. Furthermore, LLMs struggled to capture temporal dependencies within time series data and did not enhance zero-shot predictions. The authors attribute the observed performance improvements to the structure combining patching operations with attention modules. This study suggests that, at least for small datasets, the current approach of applying LLMs to time series prediction tasks does not yield performance benefits.

\section{Impact of Dataset Size on the Performance of LLM-based Models vs. SOTA Models.}\label{sec:appendix_3}

\begin{table*}[htbp]
\centering
\caption{Meaning of Symbols in the Formula}
\scalebox{1.1}{
\begin{tabular}{|c|p{10cm}|}
\hline
Symbol & Meaning \\ \hline
$ N $ & Model parameter count, representing the number of subregions into which the model divides the intrinsic space \( M \). It reflects model complexity, typically proportional to the width or number of layers in the model. \\ \hline
$ D $ & Dataset size, influencing the prediction loss in scaling laws. \\ \hline
$ d_f(H) $ & Feature dimension of the intrinsic space \( M(H) \) corresponding to the look-back horizon \( H \), representing the effective information dimension of time series data. \\ \hline
$ d_I(H) $ & Intrinsic dimension of the intrinsic space \( M(H) \) corresponding to the look-back horizon \( H \), reflecting the effective information complexity of a time series segment, typically linear with respect to \( H \). \\ \hline
$ d_I^* $ & Optimal intrinsic dimension, corresponding to the intrinsic dimension \( d_I(H) \) of the look-back horizon \( H \) that optimizes prediction performance for a given dataset size and model configuration. \\ \hline
$ \mathcal{W}(\cdot) $ & Lambert W function, used to solve equations involving products of exponential and polynomial terms, particularly in deriving the optimal look-back horizon in scaling laws. \\ \hline
$ \alpha_Z $ & Constant less than 1, describing the Zipf's law distribution of eigenvalues in the intrinsic space, representing the power exponent of the ratio of consecutive eigenvalues, affecting the decay rate of intrinsic dimensions. \\ \hline
$ C_0 $ & Constant in the scaling law loss function, dependent on parameters such as \( K_1 \), \( K_2 \), \( \eta \), and \( \lambda_0 \) (exact form to be confirmed, possibly \( \frac{K_1^2 (1 - \eta) \lambda_0}{K_2^2} \)). \\ \hline
$ K_1 $ & Constant representing the first-order Lipschitz constant of the prediction function, quantifying the smoothness of the model in the intrinsic space. \\ \hline
$ K_2 $ & Constant representing the second-order Lipschitz constant of the prediction function, quantifying the smoothness of the model in the intrinsic space. \\ \hline
$ \eta $ & Hyperparameter in the range [0, 1], controlling the weight of the noise term's contribution to the prediction loss, affecting the impact of noise variance in the Bayesian loss. \\ \hline
$ \lambda_0 $ & Base eigenvalue in the intrinsic space's eigenvalue distribution, appearing in Zipf's law (\( \lambda_i \approx \lambda_0 i^{-1/\sigma_0} \)), describing the importance distribution of intrinsic dimensions in time series data. \\ \hline
$ \sigma_M^2 $ & Noise variance of the data distribution in the intrinsic space \( M \), reflecting the randomness or uncertainty in time series data, affecting the noise term in the Bayesian loss. \\ \hline
$ S $ & Forecasting horizon, representing the number of future time steps to predict in the time series forecasting task. \\ \hline
\end{tabular}
}
\label{tab:symbol_meanings}
\end{table*}

In this section, we explore how dataset size influences the performance differences between models based on LLMs and state-of-the-art (SOTA) models. Our analysis builds on theoretical insights and notation established in prior work~\cite{shi2024scaling}. The meanings of key symbols used in the following derivations are summarized in Table~\ref{tab:symbol_meanings}.

The loss in time series forecasting models can be decomposed into two components: the Bayesian loss, representing the capability of the optimal Bayesian model, and the approximation loss, reflecting the model's ability to approximate the Bayesian estimation.
\[
L = L_{\text{Bayesian}} + L_{\text{approx}}
\]

The Bayesian loss for time series forecasting is given by:
\begin{equation}
L_{\text{Bayesian}} \approx K_1^2 (1 - \eta) \frac{\lambda_0}{(\alpha_Z - 1) d_I(H)^{\alpha_Z - 1}} + \eta \sigma_M^2 S^2 d_I(S)
\end{equation}
This term is independent of model architecture and dataset size \( D \). Since the forecasting horizon \( S \) is fixed in our experiments, the noise term \( \eta \sigma_M^2 S^2 d_I(S) \) remains constant across models, ensuring the Bayesian loss is consistent for different models.

The approximation loss is derived under the piecewise linear hypothesis, where the model divides the intrinsic space \( M \) into \( N \) subregions, performing independent linear predictions in each. Here, \( N \) represents the model’s parameter count, reflecting its complexity. The approximation loss arises from two factors: model constraints (limited by \( N \)) and dataset limitations (influenced by \( D \)).

Combining the Bayesian and approximation losses, the total loss is expressed as:
\begin{equation}
\text{loss} \approx K_2^2 \frac{d^2 N^{-\frac{4}{d}}}{4\pi^2} + K_1^2 (1 - \eta) \frac{\lambda_0}{\alpha_Z - 1} \frac{1}{d_I(H)^{\alpha_Z - 1}} + \frac{N d_I(H)}{D} \sigma_M^2 S^2 d_I(S)
\end{equation}
For detailed derivations, refer to Appendix B.3 of the literature.

When the dataset size \( D \) is small (i.e., \( N/D \) is large), the total loss is dominated by the third term, \( \frac{N d_I(H)}{D} \sigma_M^2 S^2 d_I(S) \), which is driven by the noise variance \( \sigma_M^2 \) in the intrinsic space and amplified by the ratio \( N/D \). This results in small datasets (e.g., ETTH1, ETTH2, ETTM1, ETTM2) where LLM-based models, with smaller \( N \), may achieve comparable or lower loss than SOTA models. The larger \( N \) in SOTA models exacerbates overfitting to noise, increasing their loss.

As the dataset size \( D \) grows, the noise term \( \frac{N d_I(H)}{D} \sigma_M^2 S^2 d_I(S) \) diminishes due to the inverse dependence on \( D \), and the first term, \( K_2^2 \frac{d^2 N^{-\frac{4}{d}}}{4\pi^2} \), becomes dominant, reflecting the model’s ability to partition the intrinsic space. Models with larger \( N \), such as SOTA models, achieve lower loss by better capturing the intrinsic space’s eigenvalue distribution. In contrast, LLM-based models, with smaller \( N \), exhibit higher loss on large datasets (e.g., Traffic), leading to an increased relative loss compared to SOTA models. This highlights the lower parameter efficiency of LLM-based models for time series forecasting.

Many existing works enhance LLM-based model performance by incorporating task-specific components (e.g., adapters or fine-tuned layers), inadvertently increasing the model’s capacity to partition the intrinsic space (akin to increasing \( N \)). However, LLM-based models fundamentally exhibit underfitting characteristics due to their limited parameter efficiency, particularly on large datasets where capturing complex patterns is critical.

Through theoretical and experimental analysis, it is demonstrated that with smaller datasets, a few-parameter Tokenizer-Detokenizer can fit the data well and even show better generalization and predictive capabilities for data with significant distribution shifts (such as ETTH2 and ETTM2). However, as dataset size \( D \) increases, the limited parameter efficiency of LLM-based models results in reduced prediction accuracy due to insufficient capacity to model the intrinsic space’s complexity.

\section{Why Does Train-TD Perform Better Than Train-B on ETTH2 and ETTM2?}\label{sec:appendix_4}

To investigate why Train-TD outperforms Train-B on datasets like ETTH2 and ETTM2, we compare their performance against a baseline linear model. This linear model retains the Encoder-Decoder structure but reduces the hidden dimension from 768 to 32 and replaces the 12-layer GPT Transformer with a single linear layer, resulting in a model with significantly reduced capacity. Trained on the same pre-training dataset and evaluated under zero-shot conditions, the results are summarized in Table~\ref{tab: Linear_Model}.

\begin{table}[!ht]
  \centering
  \caption{Comparison of Model Performances under Zero-shot Evaluation}
  \begin{tabular}{lcccccc}
    \toprule
     \multirow{2}{*}{\textbf{Dataset}} 
     & \multicolumn{2}{c}{Train-TD} & \multicolumn{2}{c}{Train-B} & \multicolumn{2}{c}{Linear Model} \\
    \cmidrule(lr){2-3}\cmidrule(lr){4-5}\cmidrule(lr){6-7}

    & MSE & MAE & MSE & MAE & MSE & MAE \\  

    \midrule
    \textbf{ETTH2} & 0.273 & 0.339 & 0.291 & 0.347 & \textbf{0.271} & \textbf{0.332} \\
    
    \midrule
    \textbf{ETTM2} & 0.209 & 0.292 & 0.220 & 0.296 & \textbf{0.195} & \textbf{0.284} \\

    \midrule
    \textbf{Traffic} & 0.639 & 0.429 & \textbf{0.577} & \textbf{0.400} & 0.708 & 0.468 \\
    
    \midrule
    \textbf{Electricity} & 0.263 & 0.360 & \textbf{0.232} & \textbf{0.336} & 0.253 & 0.352\\
    
    \midrule
    \textbf{Exchange rate} & 0.131 & \textbf{0.249} & \textbf{0.128} & 0.251 & 0.205 & 0.318 \\
    
    \bottomrule
    \label{tab: Linear_Model}
  \end{tabular}
\end{table}

The results show that on noisy datasets with significant distribution shifts, such as ETTH2 and ETTM2, the linear model outperforms both Train-TD and Train-B in zero-shot forecasting. In contrast, on larger and less noisy datasets like Traffic and Electricity, its performance is inferior to Train-B, which has higher capacity.

This observation motivates a theoretical analysis based on scaling laws~\cite{shi2024scaling}. When the number of model parameters \( N \) is much smaller than the dataset size \( D \), specifically when \( N = o\left(D^{\frac{d_f(H)}{d_I(H) + 4}}\right) \), the influence of dataset size on the optimal input horizon becomes negligible. The optimal input horizon \( d_I^* \) can be approximated as:
\begin{equation} 
d_I^* \approx \left( \frac{4}{\alpha_Z C_0^{\frac{1}{\alpha_Z}}} \right) \left( \ln N \right)^{1 + \frac{1}{\alpha_Z}}
\end{equation}
where \( \mathcal{W}(\cdot) \) is the Lambert W function (with \( \mathcal{W}(x) \approx \ln x \) for large \( x \)), and \( C_0 = \frac{K_1^2 (1 - \eta) \lambda_0}{K_2^2} \). For detailed derivations, refer to Appendix B.3.

This suggests that for models with small \( N \), such as the linear model, the optimal input horizon \( d_I^* \) is shorter, prioritizing recent data points. Train-B, with larger \( N \) due to pre-training, learns a longer optimal horizon, while Train-TD, with fewer effective parameters, adopts an intermediate horizon.

For high-noise datasets like ETTH2 and ETTM2, the optimal input horizon \( d_I^* \) is given by:
\begin{equation}
d_I^* = \left( \frac{K_1^2 (1 - \eta) \lambda_0 D}{N \sigma_M^2 S^2 d_I(S)} \right)^{\frac{1}{\alpha_Z}}
\end{equation}
This indicates that higher noise levels (\( \sigma_M^2 \)) require shorter input horizons to mitigate noise impact. On ETTH2 and ETTM2, the higher noise suggests a shorter \( d_I^* \), favoring the linear model and Train-TD, which use shorter horizons, over Train-B, which has a longer horizon due to its higher capacity. This explains their superior zero-shot performance.

To validate this hypothesis, we conduct experiments on the Electricity dataset with injected trend shift noise \( X + s_n(t - t_0) \) and Gaussian noise \( X + \epsilon, \epsilon \sim \mathcal{N}(0, \sigma^2) \), where \( s_n \) and \( \sigma \) control noise levels. Results are shown in Table~\ref{tab:mse_comparison}.

\begin{table*}[ht]
    \centering
    \caption{Zero-shot Evaluation of Train-TD and Train-BTD on the Electricity Dataset under Varying Noise Levels}
    \scalebox{1.15}{
    \begin{tabular}{ccccc|ccccc}
        \toprule
        \multicolumn{5}{c|}{Trend Noise} & \multicolumn{5}{c}{Gauss Noise} \\
        \cmidrule(lr){1-5} \cmidrule(lr){6-10}
        \multirow{2}{*}{\textbf{Noise Level $s_n$}}  & \multicolumn{2}{c}{Train-TD} & \multicolumn{2}{c|}{Train-BTD} & \multirow{2}{*}{\textbf{Noise Level $\sigma$}} & \multicolumn{2}{c}{Train-TD} & \multicolumn{2}{c}{Train-BTD} \\
        \cmidrule(lr){2-3}\cmidrule(lr){4-5} \cmidrule(lr){7-8} \cmidrule(lr){9-10}
        & MSE & MAE & MSE & MAE  & & MSE & MAE & MSE & MAE\\
        \midrule
        0.0002 & 0.267 & 0.370 & \textbf{0.205} & \textbf{0.313} & 0.1 & 0.299 & 0.396 & \textbf{0.249} & \textbf{0.358}\\
        0.0004 & 0.277 & 0.386 & \textbf{0.233} & \textbf{0.351} & 0.2 & 0.348 & 0.437 & \textbf{0.317} & \textbf{0.417}\\
        0.0006 & 0.296 & \textbf{0.410} & \textbf{0.285} & 0.411 & 0.3 & 0.441 & 0.503 & \textbf{0.424} & \textbf{0.494} \\
        0.0008 & \textbf{0.326} & \textbf{0.441} & 0.345 & 0.469 & 0.4 & \textbf{0.561} & \textbf{0.574} & 0.567 & 0.579\\
        0.001 & \textbf{0.368} & \textbf{0.480} & 0.428 & 0.534 & 0.5 & \textbf{0.703} & \textbf{0.645} & 0.720 & 0.650\\
        \bottomrule
    \end{tabular}}
    \label{tab:mse_comparison}
\end{table*}

The results confirm that as noise levels increase, Train-BTD’s performance degrades relative to Train-TD, particularly with trend shift noise. This supports the hypothesis that higher model capacity does not guarantee better zero-shot performance on noisy datasets with distribution shifts, as shorter input horizons are more effective.

\section{Hyperparameter Sensitivity and Forecast Horizon Analysis}\label{sec:appendix_6}

\begin{table*}[!ht]
  \centering
  \caption{Impact of Patch Size on Model Performance}
  \resizebox{\textwidth}{!}{
  \begin{tabular}{lcccccccccccc}
    \toprule
     \multirow{2}{*}{\textbf{Dataset}} 
     & \multicolumn{3}{c}{patch size=96} & \multicolumn{3}{c}{patch size=48}  
     & \multicolumn{3}{c}{patch size=32} & \multicolumn{3}{c}{patch size=24}\\
    \cmidrule(lr){2-4}\cmidrule(lr){5-7} \cmidrule(lr){8-10} \cmidrule(lr){11-13}

    & Train-TD & Train-B & Train-BTD & Train-TD & Train-B & Train-BTD & Train-TD & Train-B & Train-BTD & Train-TD & Train-B & Train-BTD \\  

    \midrule
    \textbf{ETTH1} & 0.476 & 0.427 & 0.417 & 0.466 & 0.422 & 0.410  
    & 0.473 & 0.423 & 0.413 & 0.462 & 0.422 & 0.408 \\
    
    \midrule
    \textbf{ETTH2} & 0.273 & 0.291 & 0.269 & 0.289 & 0.290 & 0.266  
    & 0.280 & 0.285 & 0.271 & 0.277 & 0.288 & 0.265 \\

    \midrule
    \textbf{ETTM1} & 0.542 & 0.419 & 0.434 & 0.537 & 0.416 & 0.428 
    & 0.544 & 0.421 & 0.430 & 0.547 & 0.417 & 0.433 \\
    
    \midrule
    \textbf{ETTM2} & 0.209 & 0.220 & 0.199 & 0.212 & 0.217 & 0.197 
    & 0.213 & 0.221 & 0.196 & 0.218 & 0.225 & 0.201 \\

    \midrule
    \textbf{Traffic} & 0.639 & 0.577 & 0.400 & 0.621 &0.563 & 0.392 
    & 0.617 & 0.557 & 0.395 & 0.619 & 0.558 & 0.394 \\
    
    \midrule
    \textbf{Electricity} & 0.263 & 0.232 & 0.198 & 0.255 & 0.225 & 0.194 
    & 0.264 & 0.228 & 0.192 & 0.252 & 0.222 & 0.195 \\
    
    \midrule
    \textbf{Exchange rate} & 0.131 & 0.128 & 0.119 & 0.130 & 0.126 & 0.120  & 0.130 & 0.126 & 0.117 & 0.129 & 0.127 & 0.118 \\
    
    \bottomrule
    \label{tab: Impact_of_Patch}
  \end{tabular}}
\end{table*}

To evaluate the sensitivity of our model to patch size choices, we conduct ablation studies by varying the patch size from 96 to 48, 32, and 24, while keeping all other settings unchanged. The prediction horizon is fixed at 96; when the patch size is smaller than 96, the model performs iterative forecasting. The zero-shot prediction results in terms of MSE are summarized in Table~\ref{tab: Impact_of_Patch}.

Reducing the patch size increases the temporal resolution of input representations, potentially capturing finer temporal patterns. For a fixed prediction horizon of 96, smaller patch sizes require iterative predictions, which may accumulate errors over multiple steps. As shown in Table~\ref{tab: Impact_of_Patch}, smaller patch sizes generally improve performance on datasets like ETTH1, ETTM1, Traffic, Electricity, and Exchange Rate, where finer granularity helps model complex patterns.  However, on high-noise datasets like ETTH2 and ETTM2, smaller patch sizes lead to slightly worse performance, likely due to error accumulation in iterative predictions amplifying the impact of noise or distribution shifts.

Despite these variations, the relative performance trends among Train-TD, Train-B, and Train-BTD remain consistent across patch sizes, with Train-BTD generally outperforming the others, particularly on larger datasets like Traffic and Electricity, indicating that our main conclusions are robust to moderate changes in patch size.

\begin{table*}[!ht]
  \centering
  \caption{Performance Comparison of Three Models at Extended Forecast Horizons}
  \begin{tabular}{lccccccc}
    \toprule
     \multirow{2}{*}{\textbf{Dataset}} & \multirow{2}{*}{\textbf{horizons}}
     & \multicolumn{2}{c}{Train-TD} & \multicolumn{2}{c}{Train-B} & \multicolumn{2}{c}{Train-BTD} \\
    \cmidrule(lr){3-4}\cmidrule(lr){5-6}\cmidrule(lr){7-8}

    & & MSE & MAE & MSE & MAE & MSE & MAE \\ 

        \midrule
        ETTH1 & 192 & 0.493 & 0.477 & 0.462 & 0.452 & 0.454 & 0.449 \\
              & 336 & 0.522 & 0.501 & 0.512 & 0.478 & 0.480 & 0.465 \\
              & 720 & 0.643 & 0.537 & 0.551 & 0.518 & 0.519 & 0.497 \\
        \midrule
        ETTH2 & 192 & 0.318 & 0.370 & 0.346 & 0.385 & 0.321 & 0.366 \\
              & 336 & 0.370 & 0.411 & 0.371 & 0.410 & 0.367 & 0.405 \\
              & 720 & 0.380 & 0.415 & 0.395 & 0.430 & 0.375 & 0.417 \\
        \midrule
        ETTM1 & 192 & 0.595 & 0.502 & 0.485 & 0.451 & 0.497 & 0.456 \\
              & 336 & 0.642 & 0.526 & 0.561 & 0.485 & 0.586 & 0.496 \\
              & 720 & 0.716 & 0.555 & 0.691 & 0.537 & 0.692 & 0.539 \\
        \midrule
        ETTM2 & 192 & 0.263 & 0.328 & 0.278 & 0.334 & 0.256 & 0.319 \\
              & 336 & 0.319 & 0.364 & 0.328 & 0.367 & 0.316 & 0.359 \\
              & 720 & 0.390 & 0.406 & 0.414 & 0.418 & 0.387 & 0.402 \\
        \midrule
        Traffic & 192 & 0.672 & 0.445 & 0.605 & 0.412 & 0.564 & 0.409 \\
                & 336 & 0.708 & 0.461 & 0.641 & 0.428 & 0.627 & 0.443 \\
                & 720 & 0.855 & 0.547 & 0.791 & 0.498 & 0.736 & 0.469 \\
        \bottomrule
    \end{tabular}
    \label{tab: performance_comparison}
\end{table*}

We also further evaluated the performance of the three models under extended forecast horizons of 192, 336, and 720 time steps. The results are summarized in Table~\ref{tab: performance_comparison}.

The relative performance ranking among the models remains largely consistent with that observed at the 96-step horizon. Specifically, Train-B and Train-BTD generally outperform Train-TD under shorter forecasting horizons. However, as the forecast horizon increases, the accumulation of iterative prediction errors becomes more pronounced, leading to a gradual degradation in overall performance for all models.

Notably, the performance gap between Train-TD and the other two models narrows with increasing horizon length. In several cases, particularly on noisy datasets such as ETTH2 and ETTM2, Train-B and Train-BTD even approach or fall below the performance of Train-TD at the longest horizon (720 steps). This suggests that while pre-training with more parameters enhances short-term predictive accuracy, its benefits diminish over longer sequences due to error propagation in iterative forecasting.

\section{Zero-shot Evaluation on Large-Scale Public Benchmarks}\label{sec:appendix_7}

\begin{figure*}[ht] 
\centering
    \includegraphics[scale=0.35]{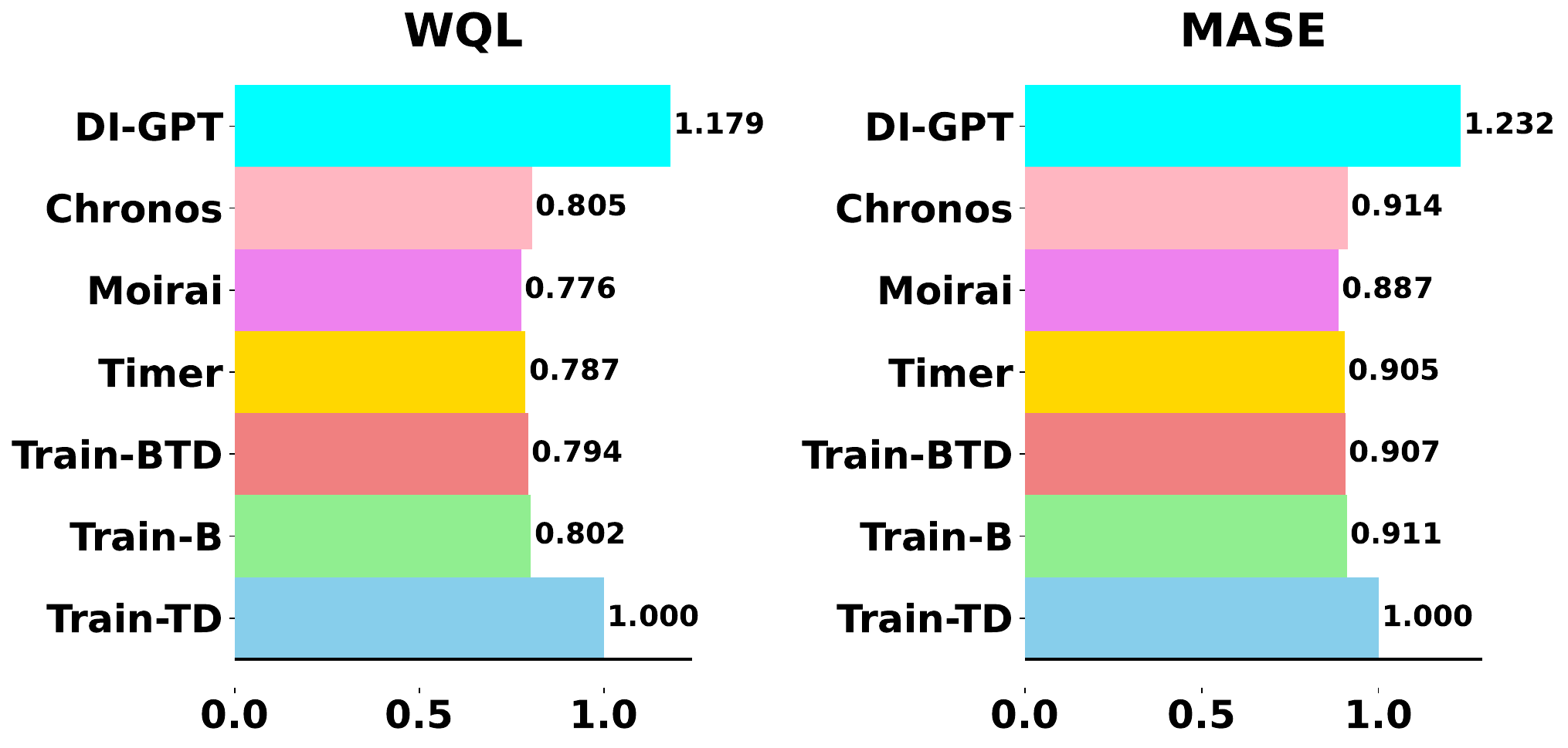}
    \Description{}
    \vspace{-1ex}
    \caption{WQL and MASE Scores from Zero-shot Evaluation on the Open Leaderboard}
    \label{f:wql}
    \vspace{-1ex}
\end{figure*}

To address recent concerns about the limited size and diversity of commonly used test sets in time series forecasting research, we further evaluate the zero-shot performance of \textbf{Train-TD}, \textbf{Train-B}, and \textbf{Train-BTD} on a large-scale public benchmark~\cite{shchur2025fev}. Figure~\ref{f:wql} presents the Weighted Quantile Loss (WQL) and Mean Absolute Scaled Error (MASE) scores, computed using the official evaluation code provided at \url{https://github.com/autogluon/fev/tree/main}. These metrics offer a more comprehensive and statistically robust assessment compared to traditional error measures. The results confirm that the relative performance of the models remains consistent with our earlier findings based on MSE and MAE. This indicates that our conclusions are not only valid within controlled experimental settings but also generalize well to larger, real-world datasets.

\section{Alternative mechanisms for leveraging semantic knowledge from pre-trained representations}\label{sec:appendix_8}

In addition to the cross-attention approach, some prior work has explored directly mapping time series patches into the vocabulary space of LLMs by replacing each patch token with a corresponding token from the LLM’s pre-defined vocabulary~\cite{cao2023tempo, pan2024textbf}. Following this paradigm, we replace the cross-attention mechanism with a direct token replacement strategy. Specifically, each time series patch is encoded as the most similar token from the LLM vocabulary, based on cosine similarity — a process conceptually similar to vector quantization. All models are then pre-trained under this new scheme.

\begin{table}[!ht]
  \centering
  \caption{Performance Comparison Between Cross-Attention and Vocabulary Replacement Strategies}
  \begin{tabular}{lcccc}
    \toprule
     \multirow{2}{*}{\textbf{Dataset}} 
     & \multicolumn{2}{c}{Cross-Attn} & \multicolumn{2}{c}{Vocab-Replace}  \\
    \cmidrule(lr){2-3}\cmidrule(lr){4-5}

    & MSE & MAE & MSE & MAE  \\  

    \midrule
    \textbf{ETTH1} & \textbf{0.782} & \textbf{0.608} & 0.800 & 0.612  \\
    
    \midrule
    \textbf{ETTH2} & \textbf{0.373} & \textbf{0.416} & 0.380 & 0.421  \\

    \midrule
    \textbf{ETTM1} & \textbf{0.731} & \textbf{0.566} & 0.747 & 0.570  \\
    
    \midrule
    \textbf{ETTM2} & \textbf{0.318} & \textbf{0.371} & 0.324 & 0.375 \\

    \midrule
    \textbf{Traffic} & \textbf{1.465} & \textbf{0.825} & 1.502 & 0.840 \\
    
    \midrule
    \textbf{Electricity} & \textbf{0.896} & \textbf{0.775} & 0.918 & 0.780 \\
    
    \midrule
    \textbf{Exchange rate} & \textbf{0.540} & \textbf{0.558} & 0.551 & 0.563  \\
    
    \bottomrule
    \label{tab: Cross_Atten_Vocab_Replacement}
  \end{tabular}
\end{table}

The results of these experiments are summarized in Table~\ref{tab: Cross_Atten_Vocab_Replacement}, showing that the direct token replacement strategy does not outperform the cross-attention method in zero-shot forecasting tasks. In fact, neither approach demonstrated significant effectiveness in our experimental setup. To better understand how time series data are represented under this paradigm, we examine several examples of the resulting token sequences:

\begin{itemize}
\item \textcolor{purple}{\texttt{J A will A his ill J}}
\item \textcolor{purple}{\texttt{le le st t le le ill}}
\item \textcolor{purple}{\texttt{k at her P I her I}}
\item \textcolor{purple}{\texttt{\textbackslash x0b \textbackslash x0b \textbackslash x0b 9 -- \textbackslash x0b am}}
\item \textcolor{purple}{\texttt{H L L L L L se}}
\end{itemize}

These tokenized representations appear as random noise rather than coherent linguistic patterns, indicating that simply aligning time series tokens with the semantic structure of LLM vocabularies is insufficient. The rich semantic knowledge embedded in pre-trained LLMs cannot be effectively exploited through such a direct mapping. Our findings suggest that more sophisticated mechanisms are required to bridge the gap between continuous time series data and discrete language representations. Both the cross-attention mechanism and direct token replacement strategies evaluated here did not prove effective in our experiments.

\end{document}